\documentclass[journal]{IEEEtran}

\ifCLASSINFOpdf
\else
\fi

\usepackage[compact]{titlesec}
\usepackage{subfig}
\usepackage{graphicx} % Required for inserting images
\usepackage[noend]{algorithm2e}
\usepackage{algorithmic}
\usepackage{setspace}
\usepackage{url}
\usepackage{amsmath,amsthm}
\usepackage{amssymb,wasysym}
\usepackage{mathrsfs}
\usepackage{cite}
\usepackage{xcolor}
\usepackage{wrapfig}
\usepackage{verbatim}
\usepackage{dsfont}
\usepackage{multirow}
\usepackage{breqn,xspace}

\ifodd 0
\newcommand{\rev}[1]{{\color{blue}#1}} %revise of the text
\newcommand{\newrev}[1]{{\color{red}#1}} %revise of the text
\newcommand{\needrev}[1]{{\color{green}#1}} %revise of the text

\else
\newcommand{\rev}[1]{#1}
\newcommand{\newrev}[1]{#1} %revise of the text
\newcommand{\needrev}[1]{#1} %revise of the text
\fi

\newcommand{\name}{FedPipe\xspace}

\newcommand{\lora}{LoRA\xspace}
\newcommand{\FedAdp}{FedAdapter\xspace}

\begin{document}

\title{Automated Federated Pipeline for Parameter-Efficient Fine-Tuning of Large Language Models}

\author{Zihan Fang, Zheng Lin, Zhe Chen~\IEEEmembership{Member,~IEEE}, Xianhao Chen,~\IEEEmembership{Member,~IEEE}, \\
Yue Gao~\IEEEmembership{Fellow,~IEEE}, and Yuguang Fang~\IEEEmembership{Fellow,~IEEE}

% \thanks{Z. Fang was with the School of Computer Science, Fudan University, Shanghai, and the work was partially done when Z. Fang is with the Department of Computer Science, City University of Hong Kong, Kowloon, Hong Kong SAR (e-mail: zhfang19@fudan.edu.cn).}

\thanks{Z. Fang, Z. Chen and Y. Gao are with the School of Computer Science, Fudan University, Shanghai 200438, China (e-mail: zhechen@fudan.edu.cn; gao\_yue@fudan.edu.cn). Z. Fang is also with the Department of Computer Science, City University of Hong Kong, Kowloon, Hong Kong SAR, China (e-mail: zihanfang3@cityu.edu.hk).}
\thanks{Z. Lin and X. Chen are with the Department of Electrical and Electronic Engineering, University of Hong Kong, Pok Fu Lam, Hong Kong, China (e-mail: linzheng@eee.hku.hk; xchen@eee.hku.hk).}
\thanks{Yuguang Fang is with the Department of Computer Science, City University of Hong Kong, Kowloon, Hong Kong SAR, China (e-mail: my.fang@cityu.edu.hk).}
\thanks{\textit{(Corresponding author: Yue Gao)}}
}

\markboth{Journal of \LaTeX\ Class Files,~Vol.~14, No.~8, August~2015}%
{Shell \MakeLowercase{\textit{et al.}}: Bare Advanced Demo of IEEEtran.cls for IEEE Computer Society Journals}

% make the title area
\maketitle

\begin{abstract}
Recently, there has been a surge in the development of advanced intelligent generative content (AIGC), especially large language models (LLMs). For many downstream tasks, it is necessary to fine-tune LLMs using private data.  
While federated learning offers a promising privacy-preserving solution to LLM fine-tuning, the substantial size of an LLM, combined with high computational and communication demands, makes it hard to apply to handle downstream tasks. 
More importantly, private edge servers often possess varying computing and network resources in real-world scenarios, introducing additional complexities to LLM fine-tuning. 
To tackle these problems, we design and implement an automated federated pipeline, named \name, to fine-tune LLMs on heterogeneous edge servers with minimal training cost and no additional inference latency. \name firstly identifies the weights to be fine-tuned based on their contributions to the LLM training. It then configures a low-rank adapter for each selected weight within the resource constraints of the edge server and aggregates these local adapters from all edge servers to fine-tune the whole LLM. Finally, it appropriately quantizes the parameters of LLM to reduce memory space \newrev{according to the requirements of edge servers.
% \needrev{based on heterogeneous computing and network resources.
Extensive experiments demonstrate that \name expedites model training and achieves higher accuracy than the state-of-the-art benchmarks.}

% AI applications powered by deep learning inference are increasingly run natively on edge servers to provide a better interactive user experience. This often necessitates fitting a model originally designed and trained in the cloud to edge servers with a range of hardware capabilities, which so far has relied on time-consuming manual effort.    
\end{abstract}


\IEEEpeerreviewmaketitle

\section{Introduction}
% Motivation: 
% 1. use lora to advoid additional inference latency
% 2. how to pluggin lora and aggregation based on the heterogeneous computing environment
% 3. use quntanzation lora to greatly reduce the number of trainable parameters (communication and computation joint optimization)

% Architecture: auto-pipeline
% 1. Use the contribution of activation to find which layer to fine-tune the pretrained model 
% 2. Resource aware configuration - determine the number of layers to train based on activation, and the rank of LoRA to use based on the computing resource of each client
% 3. Heterogeneous FL aggregation
% 4. Quantize parameters of model to reduce memory size

We are now witnessing the tremendous success of advanced intelligent generative content~(AIGC)~\cite{lin2023pushing}, especially large language models~(LLMs), such as OpenGPT~\cite{radford2019language, brown2020language}, LLaMa~\cite{touvron2023llama}, and Palm~\cite{chowdhery2023palm}, in our daily life.  Due to the huge number of model parameters~(e.g., OpenGPT3 has 175~\!B parameters), those LLMs deliver outstanding performance on various natural language processing~(NLP) applications, especially, questions and answers~\cite{jiang2023fdapt,lin2024split,qiu2024ifvit}, sentiment analysis~\cite{englhardt2023classification, nan2021deep}, etc.  It is believed that many new exciting applications will come out in the
near future such as smart healthcare, intelligent transportation,
and image classification~\cite{peng2024sums,fang2024ic3m,yuan2023graph,lin2022channel,yuan2024satsense,lin2022tracking}. Considering the state-of-the-art workflow of LLMs, generally, we can summarize into three phases: 1) pre-training LLMs from scratch with extensive computing resources~(e.g., training  GPT3-1.3B model requires 64 Tesla V100 GPUs for one week~\cite{yuan2022decentralized}) and tremendous text corpora~(e.g., Wikipedia); 2) fine-tuning pre-trained LLMs for various downstream tasks; and 3) deploying fine-tuned LLMs at an edge server to infer from input data.

Due to privacy concerns, even if many parties with the same downstream task demand a shared LLM, they cannot directly contribute their raw data to fine-tune it~\cite{lin2024efficient,hu2024accelerating,lin2024splitlora,lin2024adaptsfl}.
For example, several hospitals would like to fine-tune an LLM for a special disease diagnosis via their edge servers~\cite{karargyris2023federated,zhang2024satfed}, but they cannot share their patients' data. Although federated learning~(FL) is a promising solution to address this privacy issue~\cite{shin2022fedbalancer, panchal2024flow, zhang2024fedac, shuai2022balancefl, dai2023fedgp,lin2023fedsn}, it is non-trivial to fine-tune LLMs via FL due to the limited computing capability of edge servers. 
% since generally, comparing with core clouds, edge servers offer much less computing capability. 
In typical configurations, edge servers are outfitted with low-cost, commercially available GPUs, such as the NVIDIA GeForce RTX 3090-Ti. These GPUs, while powerful for many applications, are outmatched by the more robust GPUs found in \newrev{data center}
% central server 
environments~\cite{panopoulos2023exploring, cai2023federated}. Specifically, \newrev{data centers}
% central servers 
often deploy \newrev{data-center-grade} GPUs like the NVIDIA A100 and V100, characterized by significantly higher communication bandwidth~\newrev{(e.g., NVLink)}, larger graphic memory, and enhanced computing capabilities.  
This distinction in hardware resources has critical implications for computational tasks~\cite{chen2023confidant, ouyang2023harmony, ouyang2023efficient}, notably, edge servers face challenges in supporting full parameter fine-tuning of LLMs within standard FL frameworks. 
% \begin{table}[b] 
% \caption{\needrev{Performance and price comparison of 3090-Ti and A100}}
% \centering
% \begin{tabular}{ |c|c|c|}
%  \hline
%                 & 3090-Ti & A100 \\
%  \hline
%  Price           & \$2,000 & \$14,000 \\
%  FP32 Performance  & 36 TFlops   & 19 TFlops  \\
%  Tensor Cores   & 336   & 432 \\
%  \hline
% \end{tabular}
% \end{table}\label{tab:per_gpu}

\begin{figure}[t] 
\centering
\includegraphics[width=\linewidth]{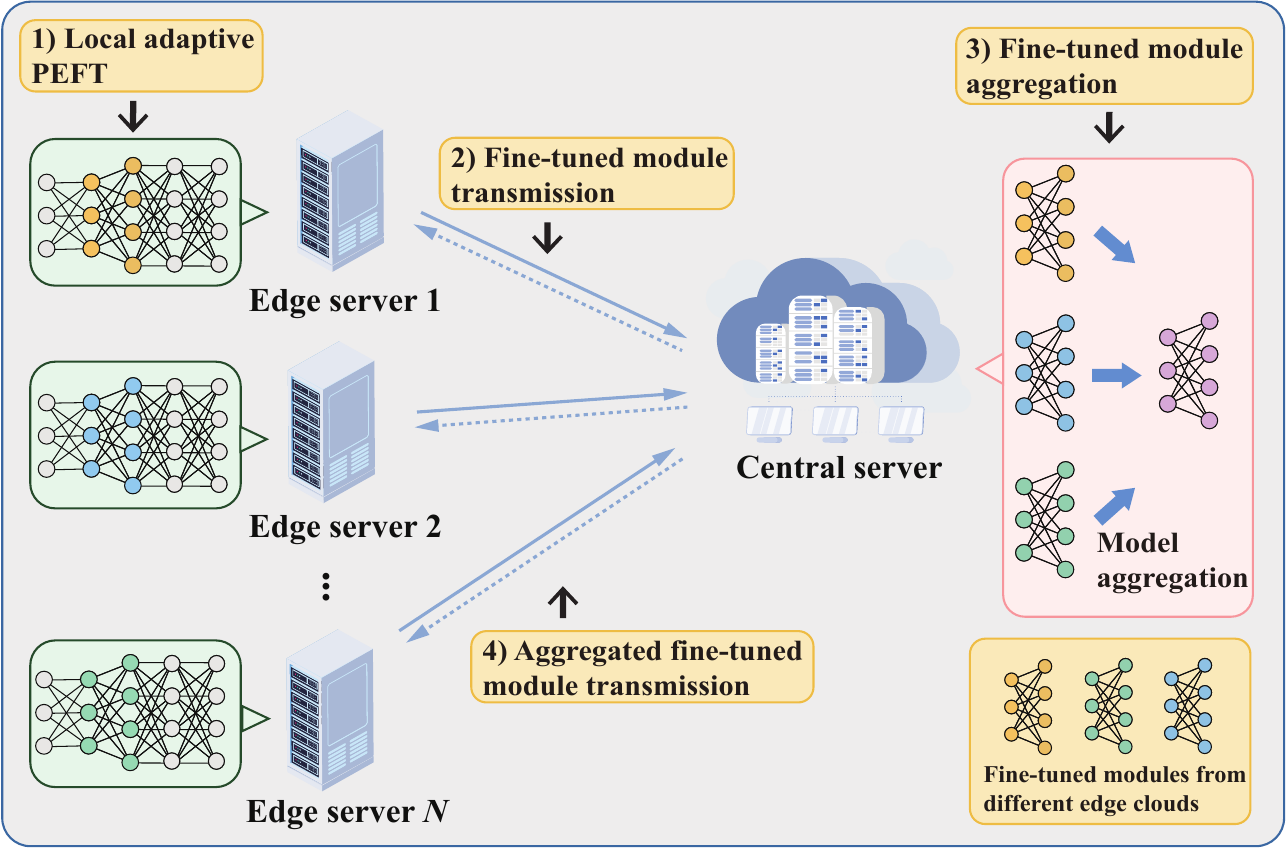}
% \vspace{-2mm}
\caption{A scenario of LLM FL \newrev{via PEFT} with edge servers.}
% \label{Scenario}
\label{fig:llm_fl_teaser}
\vspace{-2ex}
\end{figure}

Fortunately, the integration of FL and Parameter-Efficient Fine-Tuning (PEFT) may solve the above challenges. Recent research efforts~\cite{houlsby2019parameter, pfeiffer2020adapterfusion, hu2021lora, karimi2021compacter, sung2021training, zhang2023gpt, hu2023llm, sheng2023s} have explored PEFT methods, \newrev{in particular,} \textit{ADAPTER}~\cite{houlsby2019parameter, pfeiffer2020adapterfusion, karimi2021compacter} and \textit{LoRA}~\cite{hu2021lora, sheng2023s}, to mitigate the challenges of intensive computational resources required by LLM fine-tuning. Fig.~\ref{fig:llm_fl_teaser} illustrates how to integrate PEFT with FL for LLMs. The central server distributes a pre-trained LLM to edge servers~(participants), and then edge servers perform local training using PEFT methods. After training, the central server aggregates these updates and redistributes the enhanced LLM to edge servers for subsequent training rounds. This iteration continues until the LLM achieves a predefined accuracy threshold. Recent work, \textit{FedAdapter}~\cite{cai2023efficient}, adopts ADAPTER to achieve efficient FL for NLP downstream fine-tuning, but mainly focuses on standard language model, such as BERT~\cite{kenton2019bert}. A few research works~\cite{liu2023federated, che2023federated} study LLM FL with PEFT, but they always assume every edge server with homogeneous resources~(i.e., computing power and memory). \newrev{Consequently}, these frameworks are not well adapted to heterogeneous environments and cause deterioration in training performance. In practice, it is still non-trivial to integrate PEFT with FL in LLMs for downstream tasks.

In this paper, we propose and implement an automated \underline{fed}erated \underline{pipe}line~(\name) for LLM FL fine-tuning. 
Unlike existing \newrev{PEFT} methods, \name does not aim to introduce new PEFT techniques. Instead, it serves as a foundational framework to enhance any existing PEFT method. This is achieved by systematically decoupling, abstracting, and automating the operations and parameters of these methods. As a result, application and algorithm developers can efficiently outsource the implementation intricacies of LLM FL fine-tuning, allowing them to concentrate on their main services or model design.
%
% for purpose of reducing large communication overhead, instead of local LLMs, local adapter of each edge server needs to transmit for model aggregation
Despite the promising potential of \name, how to implement this framework poses significant challenges. 
First, given the varying computing resources across edge servers, this heterogeneity causes a pronounced straggler problem in FL~\cite{9887795, 9093123, zhang2023timelyfl, horvath2021fjord, ouyang2022clusterfl,lyu2023optimal}. For instance, two edge servers with slightly different GPU memories (e.g., 22\!~GB vs. 24\!~GB) can experience a training time discrepancy of several hours per communication round in FL.
% \name must address the straggler problem by adapting its adapters to these heterogeneous resources during training. 
Second, to offer better fine-tuning performance, \name enables each edge server to identify and prioritize important weights for adapter construction. However, assessing and modeling the importance of these weights remains an open problem.
Finally, each adapter within \name must align well with the computing budget of its respective edge server, which demands careful adjustment of model parameters. 
%Last but not the least, since GPU memory of edge server is limited, it is hard to put the whole LLM in the GPU, and train the corresponding adapter. 
These challenges underscore the complexity and necessity of careful design of \name.

% in order to reduce large communication overhead, instead of local LLMs, local adapter of each edge server needs to transmit for model aggregation, but   

% First, for LLM FL fine-tuning, in order to reduce large communication and computation overhead, according to data diversity on edge servers, fine-tuning structures of an LLM are different. How to figure out the best structure to train in an LLM based on a dataset of an edge server is very hard. Second, due to heterogeneous computing resources in edge servers during a training phase, to ensure all fine-tuning structures training synchronization, those structures have to adapt heterogeneous resources of edge servers via adjusting model parameters. Third, for most of edge servers, their memory sizes are usually limited, and how to further accelerate training in that case is an open problem. 
% We need to identify the importance of each layer, and combine them into a fine-tuning structure based on a dataset of an edge server. 
% although we leverage PEFT methods to reduce training time, the whole FL procedure including distribution, training, collection, and aggregation still encounters large communication and computation overhead. 

% \name's high-level design 

In this paper, we first conduct pilot measurements to analyze the above challenges and acquire the corresponding insights to guide the system design of \name. Then, we formulate the automated LLM FL fine-tuning pipeline as a Mixed-Integer Linear Programming~(MILP) optimization problem. To solve the MILP problem, \newrev{we propose a two-stage method, comprised of trainable weights identification method and fast search algorithm.} The adaptive identification method of important weights is \newrev{used to} figure out the best adapter of an LLM to fine-tune 
% weight matrices 
based on their importance score for each edge server. \newrev{This method parameterizes incremental updates~(i.e., trainable weights) using singular value decomposition~(SVD) to effectively prune unimportant updates with the lower singular values, leading to the efficient adapter with fewer trainable weights.
}
% the singular values of unimportant updates, leading to a reduction parameter budget. 
According to the adapters of edge servers, by considering heterogeneous computing resources of edge servers, we design a fast search algorithm \newrev{to select the best set of rank and batch size for local training at each edge server.}
% for \newrev{computing budgets alignment.} 
% adapters configuration. 
% The algorithm selects the best set of rank and batch size for local training at each edge server. 
Finally, 
% to ensure LLM FL training for edge servers, 
\newrev{to consider diverse memory budgets for edge servers,}
and further accelerate their training speed, we leverage a simple quantization-agnostic backward pass to adaptively achieve low-precision LLM weights from a customized black-box quantization module. Moreover, different from existing LLM FL aggregation approaches, we design a lightweight partial weights aggregation approach by only aggregating local adapters to improve communication efficiency. 

We summarize our main contributions of \name as follows.
\begin{itemize}
    \item \name is the first general automated federated pipeline of fine-tuning LLMs for \newrev{all kinds of} downstream tasks.
    \item We formulate the pipeline as an MILP optimization problem and develop an efficient algorithm to solve it.
    \item By taking heterogeneous computing resources at edge servers into consideration, we develop an effective method to find different low-rank adapter structures for heterogeneous edge servers.
    % , which is needed in efficiently solving our MILP problem. 
    \item To further reduce the computational complexity in solving our MILP at edge servers, we also design a fast search algorithm to dynamically identify important parameters for local adapter training. 
    \item Furthermore, according to those parameters and \newrev{memory budgets of edge servers, we quantize the local models into different quantization bits} to increase training efficiency.
    \item We design a partial weight aggregation approach to aggregate all adapters without increasing communication overhead.
\end{itemize}

The rest of this paper is organized as follows. Section~\ref{sec:bg} introduces the background and motivation
% and Section~\ref{sec:analysis} gives the design analysis of \name, 
and the system design of \name is presented in Section~\ref{sec:design}. Sections~\ref{sec:impl} and~\ref{sec:eval} respectively elaborate \name's implementation and report the extensive evaluations of \name. Section~\ref{sec:rw} briefly discusses the related works, followed by the conclusion in Section~\ref{sec:con}.

\section{Background and Motivation} \label{sec:bg}
Before we delve into detailed design analysis and details, we briefly introduce the background of PEFT for LLM FL. Then, to better motivate the design of \name, we provide extensive pilot measurement studies to elaborate the design challenges of FL on LLMs.

\subsection{Injecting PEFT into Federated Learning on LLMs} \label{ssec:bg:peft}

\begin{figure}[t]
\centering
\includegraphics[width=0.6\linewidth]{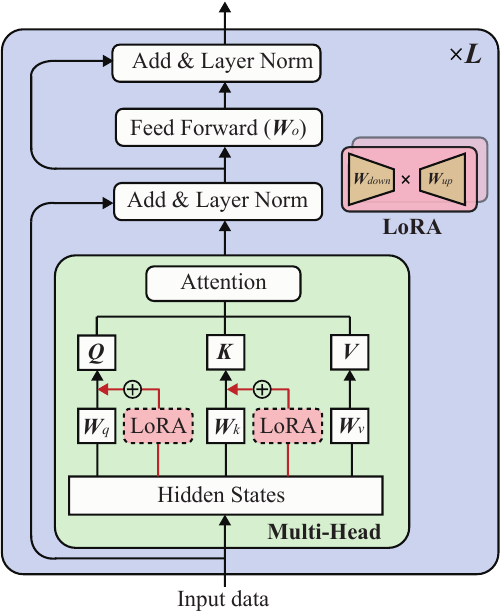}
\caption{An illustration of LoRA method.}
\label{fig:eg_fine_tune}
\vspace{-2ex}
\end{figure}

Due to the constrained computing resources at an edge server~(i.e., commercial GPUs, such as NVIDIA GeForce RTX 3090-Ti) and the large number of parameters in an LLM, full parameter fine-tuning (i.e., initializing the model with pre-trained weights, updating all parameters, and generating separate instances for various tasks) at each edge server leads to unacceptable computing and networking latency in both model training and aggregation. Table~\ref{tab:popular_llm} illustrates the number of parameters and sizes of several popular LLMs. Obviously, the smallest model, GPT-2~\cite{radford2019language} with 774\!~M parameters, is over 7 times larger than the conventional language model~(e.g., BERT~\cite{kenton2019bert}) with 110\!~M parameters. 
% Therefore, to fully fine-tune those LLMs via commercial GPUs~(e.g., RTX 3090）in edge servers and collect their local LLMs for model aggregation brings huge computing and networking overhead.
% To better understand the computing and communication bottlenecks in the FL for LLMs, we set up two experiments using one of the most prevalent LLMs, \needrev{LLama-2-x}. Figure~\ref{} and \ref{} show 
\begin{table}[b]
\centering
\scalebox{0.8}{
\begin{tabular}{ |c|c|c|}
 \hline
 Model &\#Trainable Parameters & Size \\
 \hline
 {BERT}     & 110\!~M  & 0.44\!~GB \\
 \hline
 {GPT-2}    & 774\!~M  & 3\!~GB \\
 \hline
 {GPT-3}    & 175\!~B  & 700\!~GB  \\
 \hline
 {LLaMA-1}  & 65\!~B   & 260\!~GB \\
 \hline
 {LLaMA-2}  & 70\!~B   & 280\!~GB \\
 \hline
\end{tabular}
}
\caption{Popular LLMs}
\label{tab:popular_llm}
\vspace{-2ex}
\end{table}

To combat the above stumbling obstacles in LLM FL, PEFT employed in downstream task training is a potential solution. Different from the full parameter fine-tuning, PEFT only fine-tunes parts of neurons to reduce trainable parameters in a model, leading to less computing cost. Recently, there are mainly two types of PEFT, ADAPTER~\cite{houlsby2019parameter, pfeiffer2020adapterfusion, karimi2021compacter} and LoRA~\cite{hu2021lora, sheng2023s}. As shown in Fig.~\ref{fig:eg_fine_tune}, ADAPTER inserts a few layers for each transformer block~\cite{houlsby2019parameter} and only trains those layers in a pre-trained model. Apparently, ADAPTER introduces extra latency in the inference phase, due to its additional layers~\cite{hu2021lora}. However, LoRA is an adapter with low-rank adaptation without additional inference latency that injects trainable rank decomposition matrices into each transformer layer, but freezes all weights of a pre-trained model. Many works~\cite{hu2021lora, hu2023llm, sheng2023s} have verified that LoRA outperforms ADAPTER, and hence in the following, we only focus on LoRA.

\begin{figure}[t]
\centering
\subfloat[Computing overhead\label{fig:llm_train:computing}]{
\includegraphics[width=0.495\linewidth]{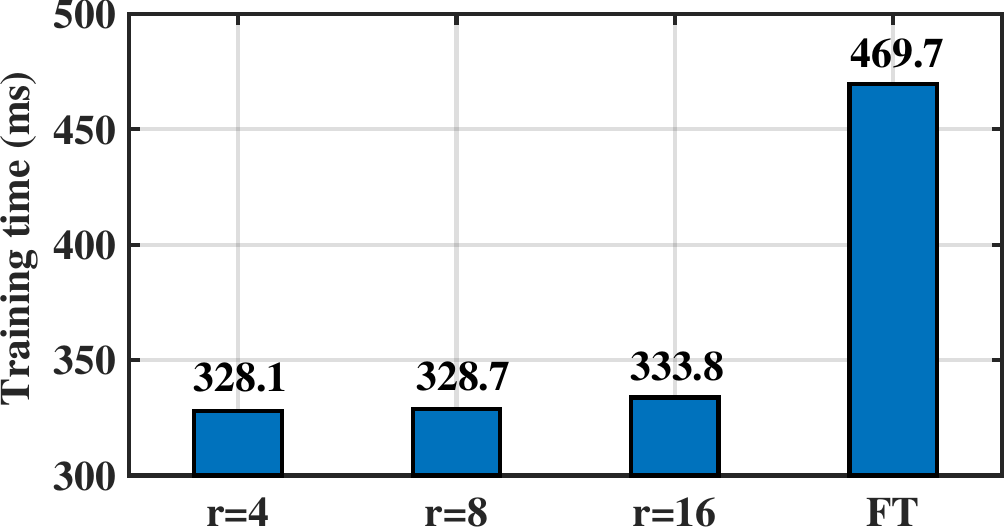}
}
\subfloat[Communication overhead\label{fig:llm_train:communication}]
{
\includegraphics[width=0.495\linewidth]{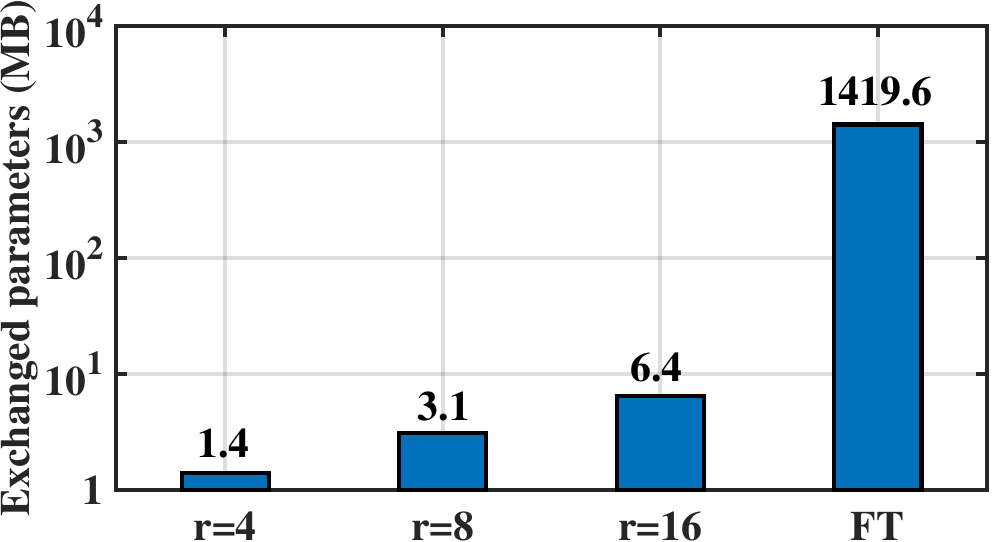}
}
\caption{The \newrev{training time, and exchanged parameters} of LoRA and full parameter fine-tuning for GPT-2, where $r$ is the rank size of LoRA adapter.}
\label{fig:llm_training}
\vspace{-2ex}
\end{figure}

To better understand the computing and communication bottlenecks in the LLM FL, we set up two experiments using one of the most prevalent LLMs, GPT-2. Fig.~\ref{fig:llm_train:computing} and \ref{fig:llm_train:communication} show the training time \newrev{for a batch of data} and exchanged parameters of full parameter fine-tuning and PEFT. Full parameter fine-tuning obtains more than 1.4 and 1000 times larger than those of PEFT in training and exchanging, respectively. Apparently, instead of the entire model fine-tuning, PEFT can significantly mitigate computing and communication overhead. However, we also meet three main challenges to integrate PEFT into LLM FL. We will use ``training'' to represent fine-tuning in the following sections.

\begin{figure}[t!]
  \centering
  \subfloat[Under-training rate\label{fig:straggler:comp}]{
    \includegraphics[width=0.505\linewidth]{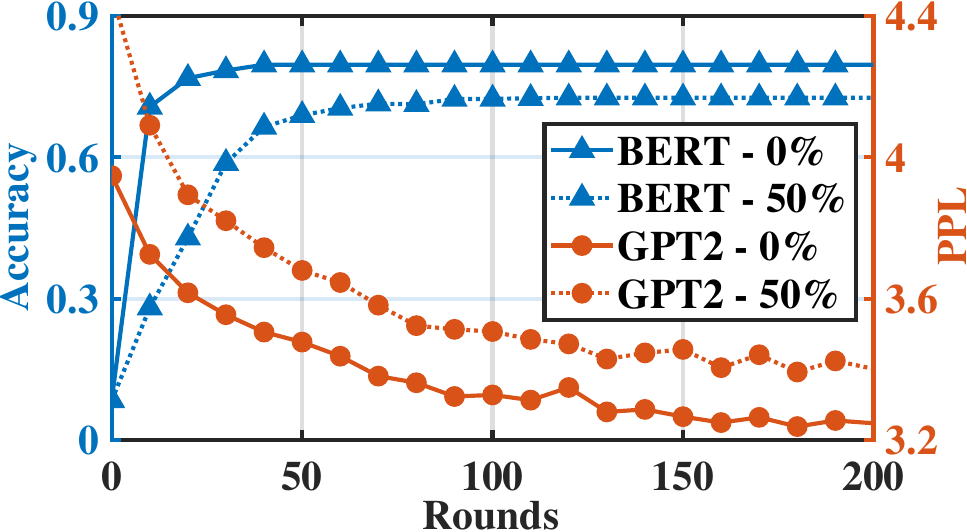}
  }
  \subfloat[Average aggregation time \label{fig:straggler:comm}]
  { \includegraphics[width=0.4850\linewidth]{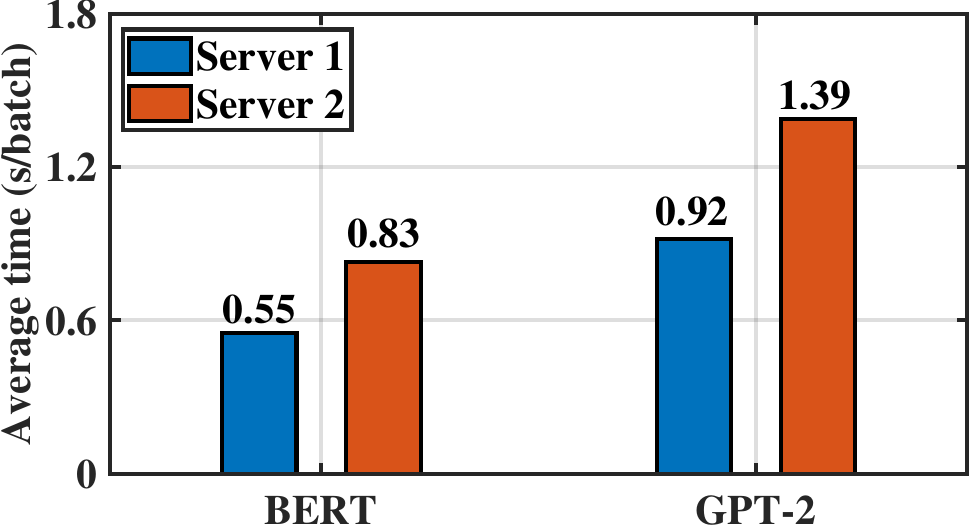}
  }
  % \vspace{-2mm}
  \caption{The straggler problem on GPT-2 and BERT.}
  \label{fig:straggler}
  \vspace{-2ex}
\end{figure}

\subsection{Severe Straggler Problem in LLM FL} \label{ssec:bg:straggler}

State-of-the-art LLM FL frameworks~\cite{liu2023federated, che2023federated} usually assume that each edge server has sufficient local resources (e.g., computing power, communication bandwidth, and memory space) to perform local model training. Therefore, a central server can easily aggregate these local models to achieve good performance. However, in practice, the available resources of various edge servers vary significantly, and the allocated resources for training may change at run-time depending on how on-demand running programs prioritize resource allocation. More importantly, due to the large parameters of LLM, the resource heterogeneity of edge servers leads to significantly different training time, and a severe straggler problem in model aggregation.

% Due to the resource heterogeneity of edge servers, model aggregation leads to under-training local models for several end devices.

To better understand the impact of heterogeneous resources on the straggler problem for LLM FL, we conduct two experiments for a standard model and LLM. We integrate LoRA into FedAvg~\cite{mcmahan2017communication},  one of the most prevalent FL frameworks, to evaluate its performance on the widely recognized LLM GPT-2 and standard model BERT for homogeneous and heterogeneous cases.
% We implement an LLM FL framework via employing FedAvg~\needrev{\cite{}}, and LoRA to evaluate its performance under both two experiments. We utilize the widely recognized LLMs GPT-2 for local model training at each edge server. 
% to evaluate its performance under three experiments, and utilize the widely recognized LLMs GPT-2 for local model training. 
We set up two PC servers (i.e., edge servers) with/without varying computing power by controlling different graphic memories of GPU~(22\!~GB and 24\!~GB), but the same ones~(24\!~GB) for homogeneous and heterogeneous cases, respectively. The result is shown in Fig.~\ref{fig:straggler}
% 10 end devices with varying computing power, communication capability and memory size, then scale their budgets to simulate the case where several end devices are under-training.
% The performance of FL paradigm for LLMs under different under-training rates caused by heterogeneous on-device resources is shown in Figure.~\ref{}, 
where the under-training rate is defined to be 
% the proportion of the under-training edge servers, which is 
the number of under-training edge servers divided by the total number of edge servers and the average aggregation time means the per-round time from the start of local model training \rev{for a batch of data} to the end of model aggregation.

It is clear to see that the straggler problem occurs in the heterogeneous case for both the standard model and LLM. However, for the heterogeneous case, even if the difference in computing capabilities between two edge servers is 10~\!TFLOPS, LLM achieves much worse performance than standard one in both under-training rate and aggregation time. 
% Apparently, the straggler problem of LLM is much heavier than that of the standard model.
Therefore, we need to develop an efficient FL fine-tuning pipeline for LLMs to accommodate heterogeneous resources on different edge servers.
% from Figure.~\ref{} and Figure.~\ref{} that the performance of FL paradigm for LLMs gradually deteriorates as ascending the under-training rate. Obviously, budget-unconstrained FedAvg can effectively aggregate local models from all end devices to achieve the fastest convergence speed and the highest accuracy. However, FedAvg, as a static FL framework, is incapable of adapting to heterogeneous on-device resources, which severely affects its performance. Therefore, we need to develop an emerging FL framework for LLMs to accommodate heterogeneous on-device resources. Apart from heterogeneous on-device resources, we also need to tackle the challenge of limited memory size, which is discussed in the following section.

% \begin{figure}
% \centering
% \includegraphics[width=\linewidth]{Figures/hetero_acc_bert.eps}
% \caption{Heterogeneous on Bert Model}
% \label{Fig1}
% \end{figure}

\begin{figure}[t!]
  \centering
  \subfloat[Quantity\label{fig:diff_config:layers}]{
    \includegraphics[width=0.505\linewidth]{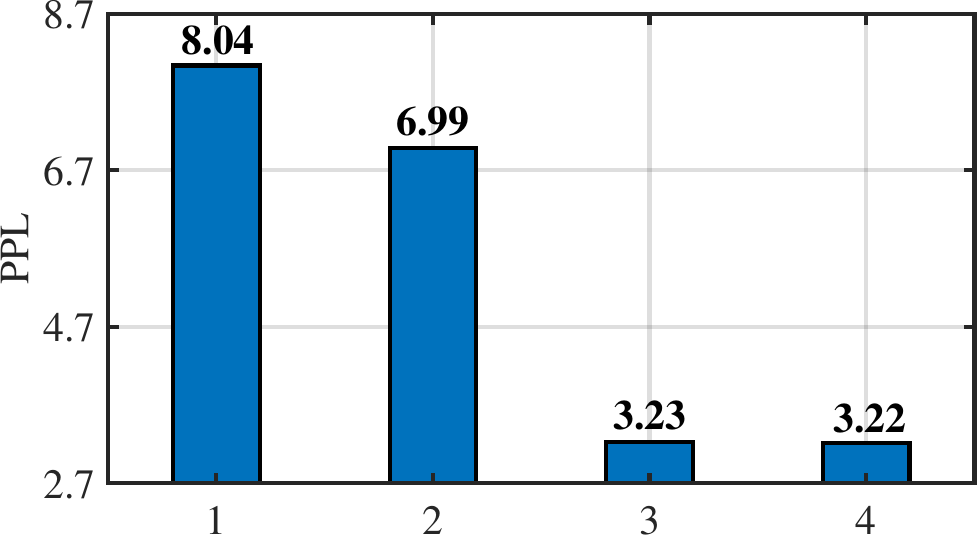}
  }
  \subfloat[Index \label{fig:diff_config:chan}]{
  \includegraphics[width=0.5\linewidth]{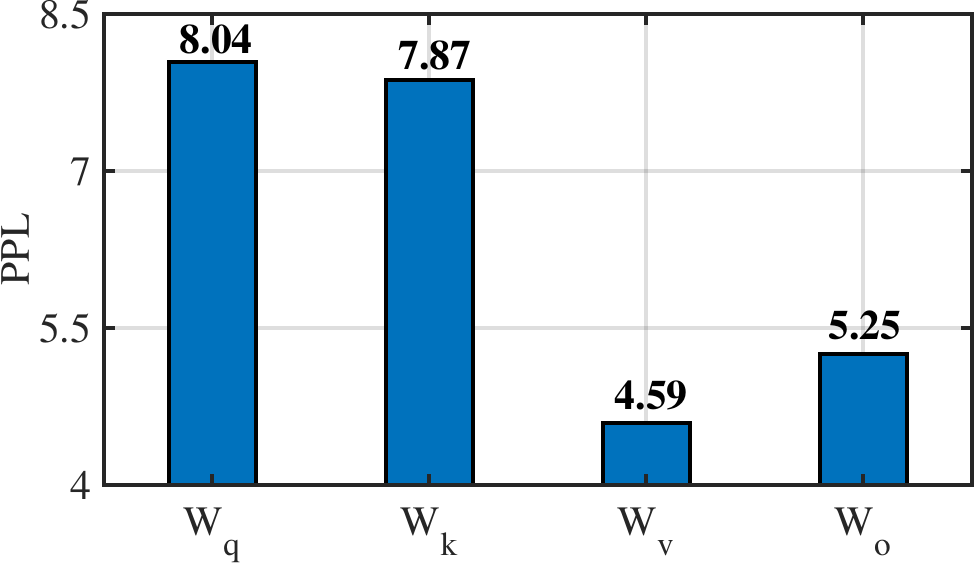}
  }
  \caption{GPT-2 training performance with different trainable weights, where the trainable weights combinations corresponding to quantities 1, 2, 3, and 4 are $\left\{ {{{\bf{W}}_q}} \right\}$, $\left\{ {{{\bf{W}}_q},{{\bf{W}}_k}} \right\}$, $\left\{ {{{\bf{W}}_q},{{\bf{W}}_k},{{\bf{W}}_v}} \right\}$, and $\left\{ {{{\bf{W}}_q},{{\bf{W}}_k},{{\bf{W}}_v},{{\bf{W}}_o}} \right\}$, respectively.}
  \label{fig:diff_config}
  \vspace{-2ex}
\end{figure}

\begin{figure}[t]
  \centering
  \subfloat[PPL\label{fig:important_ppl_20round}]
  {
    \includegraphics[width=0.50\linewidth]{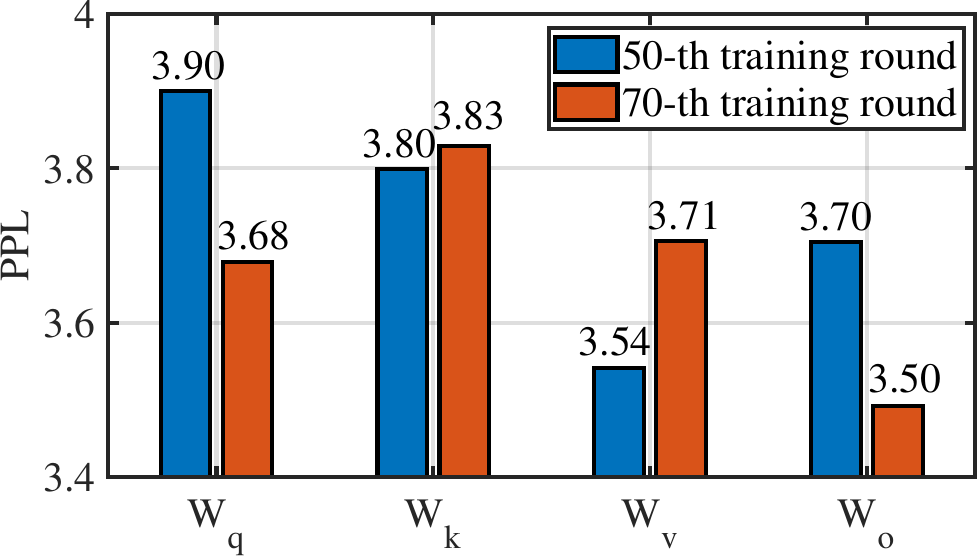}
  }
  \subfloat[STD of PPL\label{fig:important_std}]
  {
    \includegraphics[width=0.47\linewidth]{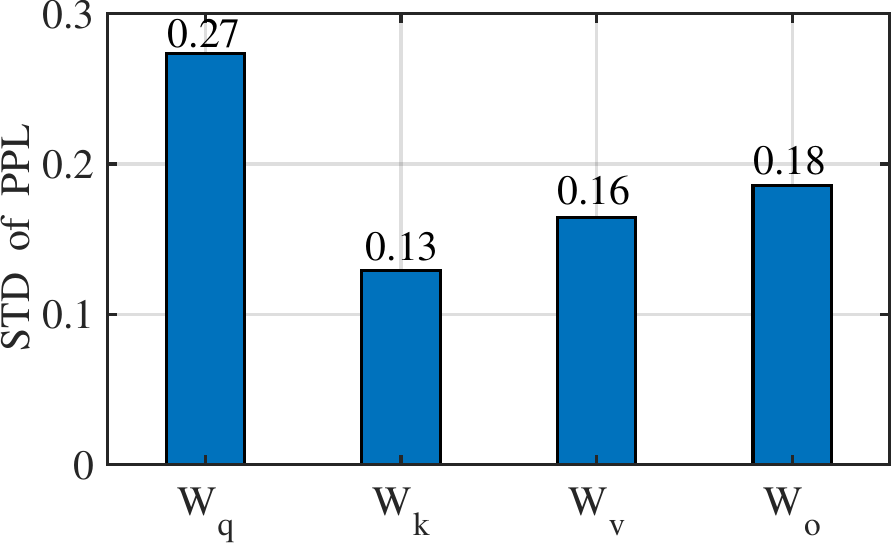}
  }
  \caption{The PPL at the $50$-th and $70$-th training rounds and the STDs of the PPL across \newrev{100} training rounds. }
  \label{fig:weight_dynamic}
  \vspace{-2ex}
\end{figure}

\subsection{Weight-Level Configuration} \label{ssec:bg:weight}

Although LoRA has reduced large computing overhead~(see Section~\ref{ssec:bg:peft}) for LLMs, due to the constrained computing capacity at an edge server, \rev{training all weights with LoRA in each layer} still poses a heavy workload. We train multiple LoRA adapter versions by increasing \rev{the number of trainable weights in each layer} shown in Fig.~\ref{fig:diff_config:layers}. \rev{The perplexity~(PPL) is used as the performance indicator to evaluate how well the model has learned the distribution of the data it was trained on, and a smaller PPL represents better performance.}
Apparently, the more weights the LoRA adapter has, the higher the computing overhead, and the better the performance required and obtained. A naive solution is to configure the number of trainable weights based on a computing budget. 

However, how to find the prior weights is a non-trivial task. We have conducted another measurement experiment to study the impact of weight indices 
% while fixing the number of layers in LoRA training. We randomly select three layer indices from the whole layer indices 
\rev{in LoRA training. We select four trainable weight vectors (i.e., $\mathbf{W}_q$, $\mathbf{W}_k$, $\mathbf{W}_v$, $\mathbf{W}_o$ shown in Fig.~\ref{fig:eg_fine_tune}) from the transformer block} to train LoRA adapters \newrev{in an round} illustrated in Fig.~\ref{fig:diff_config:chan}. It demonstrates that various weight indices have different performances. \newrev{We also plot PLLs with four trainable weight vectors in different training rounds shown in Fig.~\ref{fig:important_ppl_20round}, and calculate their standard deviations~(STDs) illustrated in Fig.~\ref{fig:important_std}. Apparently, the importance of four trainable weight vectors changes for the LoRA adapter in each training round, and hence, a one-shot measurement cannot identify their priority.}
Therefore, we need to investigate how to identify and prioritize the important trainable weights automatically for each edge server under its computing budget.

\subsection{Budget-Aware Model Parameters Alignment} \label{ssec:bg:sync}

\begin{figure}[t]
  \centering
  \subfloat[Batch size\label{fig:batch_gpt2}]
  {
    \includegraphics[width=0.505\linewidth]{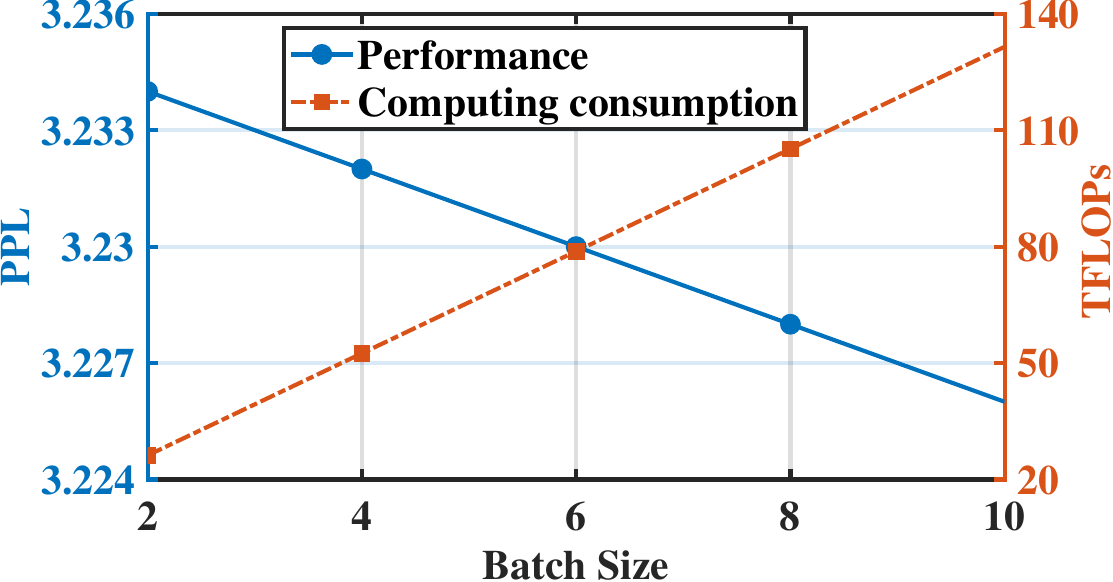}
  }
  \subfloat[Decomposition rank\label{fig:rank_gpt2}]
  {
    \includegraphics[width=0.505\linewidth]{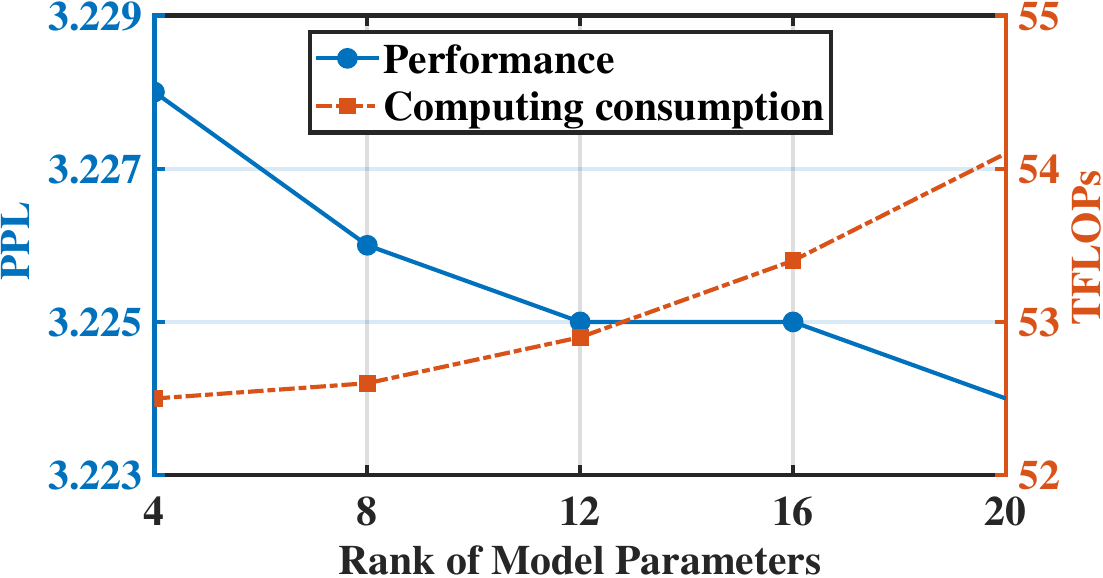}
  }
  \caption{Performance and computing consumption of GPT-2 with LoRA adapter with different training batch sizes and ranks.}
  \label{fig:batch_and_rank}
\end{figure}

While the training weights are determined, several parameters still impact LoRA training, in particular, \textit{batch size}, \textit{decomposition rank}, \textit{dropout}, \textit{alpha}, etc. We evaluate all of them and find that batch size and decomposition rank are the two most important parameters for training performance\footnote{The same results are also observed in~\cite{sheng2023s, li2024caraserve}}. To study the impact of both batch size and decomposition rank, we increase them, respectively, to train the pre-trained GPT-2 with LoRA, and the results are illustrated in Fig.~\ref{fig:batch_and_rank}. 
% It demonstrates the performance and GPU memory size of pre-trained GPT-2 with LoRA in different training batch sizes and decomposition ranks. 
It is clear to observe that a larger batch size or decomposition rank always \needrev{offers better performance but consumes more computing resources}. However, given the computing budget of an edge server, we cannot always select the largest batch size and rank. We need to find an optimal combination of batch size and rank under the computing budget for each edge server in an automated federated pipeline.

% Figure~\ref{fig:batch_and_rank} demonstrates the performance and sensitivity of pre-trained GPT-2 with LoRA in different training batch sizes and decomposition ranks. Either a larger batch size or greater decomposition rank provides outcome enhancement in several model architectures. In addition to the training performance, the batch size and rank can also affect the computing resource and memory size occupied during the training phase. Similar conclusions can be drawn as shown in Figure~\ref{fig:batch_and_rank}, indicating the possibility of performing the local models with different computing and storage capabilities in each edge server according to configuring the batch size and rank of the local model. 

% \subsection{Limited On-Device Memory Size for Deployment} \label{ssec:limited_memory}

% 存储模型参数：计算机内存用于存储模型的权重、偏置和其他参数。这些参数是训练过程中不断更新的，计算机内存必须足够大以容纳它们。大型语言模型的参数通常非常庞大，需要大量内存来存储。

% 存储训练数据：在训练期间，模型需要反复读取训练数据批次（mini-batches）。这些数据通常被加载到内存中以加速训练，因为内存中的数据可以更快地访问，而不需要频繁的磁盘读取。内存的大小必须足够以容纳当前训练批次的数据。

% 存储中间计算结果：训练过程中，模型进行大量的前向传播和反向传播计算。中间结果，如梯度、损失值和其他计算，也需要存储在内存中，以便进行参数更新和优化算法的运行。

\section{Automated Federated Pipeline Design} \label{sec:design}

\subsection{Overview of \name}
\begin{figure}[t]
\centering
\includegraphics[width=\linewidth]{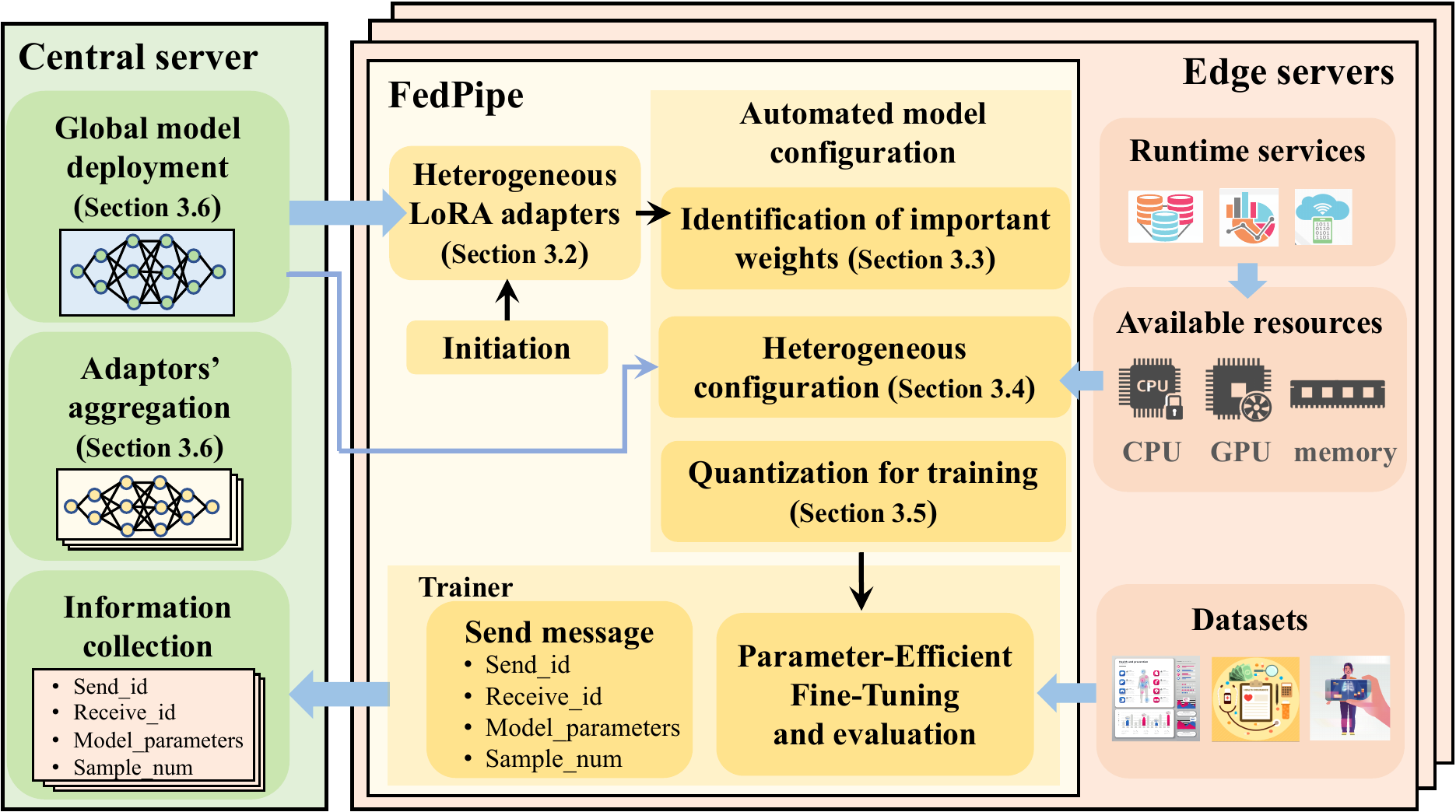}
\caption{The system overview of \name.}
\label{fig:overview}
\vspace{-2ex}
\end{figure}

To tackle the above challenges, we design \name, an automated federated pipeline shown in Fig.~\ref{fig:overview}, to determine the optimal fine-tuning model structure to improve the performance of fine-tuning LLMs, accommodating heterogeneous and resource-constrained edge servers.
% \name aims to determine the optimal fine-tuning model structure to improve the performance of fine-tuning LLMs, accommodating heterogeneous and resource-constrained edge servers. 
% As shown in Figure.~\ref{fig:overview}, \name is a unified FL framework for LLMs, enabling collaborative and efficient fine-tuning of LLMs across multiple edge servers in a privacy-preserving manner.
% In the initial step, the central server dispatches LLMs and their corresponding model parameters to participant edge servers. The LoRA adapters are automatically configured to train by \name for the edge servers. 
We first theoretically formulate an MILP to address the straggler problem under heterogeneous and constrained computing resources in LLM FL to guide \name's design (Section~\ref{ssec:lora}). To solve the problem, we propose two-level solutions to identify the important weights (Section~\ref{ssec:import_identify}) and heterogeneous adapters' configuration (Section~\ref{ssec:adp_cfg}). \newrev{We also consider heterogeneous storage memories and} quantize the LLM  based on the budget of an edge server (Section~\ref{ssec:quant}). Finally, we design an effective method to aggregate all local models with \needrev{different size of LoRA adapters} (Section~\ref{ssec:aggregate}). After that, the next training round starts. 

%[zhe] GPU memory??

% Before model training, each edge server downloads the model parameters from the central server as initial parameters. Due to the superiority of the low-rank structure for LLM fine-tuning~\cite{aghajanyan2021intrinsic, grasedyck2013literature, li2018algorithmic}, we employ the low-rank adapters to fine-tune the LLM. Additionally, we theoretically formulate an optimization problem involving batch size and \rev{LoRA rank} selection (Section~\ref{ssec:lora}), which guides \name framework design. We begin by designing an important layer identification mechanism (Section~\ref{ssec:import_identify})  to dynamically identify appropriate model layers for fine-tuning and prioritize them based on the importance level. Considering the diverse computing capabilities of edge servers, we select the optimal batch size (Section~\ref{ssec:adp_cfg}) and then adjust the fine-tuning \rev{LoRA rank} according to the importance order of the layers to enhance training performance. Apart from the computing resources, the limited memory size is also taken into consideration where \name compresses models into different quantization bits based on varying available memory sizes (Section~\ref{ssec:quant}). Finally, we propose an effective aggregation strategy to aggregate local models with different model structures and quantization bits (Section~\ref{ssec:aggregate}). After model aggregation, the aggregated fine-tuned modules are distributed to all edge servers for the next training round.

\subsection{\newrev{Modelling Heterogeneous LoRA Adapters}} \label{ssec:lora}

% LLM has gained significant attention in academia owing to its broad range of applications in daily life. However, training an LLM for a specific task directly is prohibitively expensive due to the vast datasets required for performance improvement and the substantial training time consumed by the enormous amounts of trainable parameters. While fine-tuning a pre-trained general LLM proves effective, it poses great challenges for the computing capacity and memory size of edge servers in traditional FL frameworks. To address this, an adapter has been introduced as a PEFT approach. Despite adapter layers containing only few parameters (sometimes less than 1\% of the original model), they can only be processed sequentially, which prevents hardware parallelism from accelerating training process. This limitation becomes evident in online inference where the batch size is typically as small as one, as~\cite{hu2021lora} demonstrates that the use of adapters in the GPT-2 medium model leads to a noticeable increase in inference latency.

\begin{figure}
\centering
\includegraphics[width=0.8\linewidth]{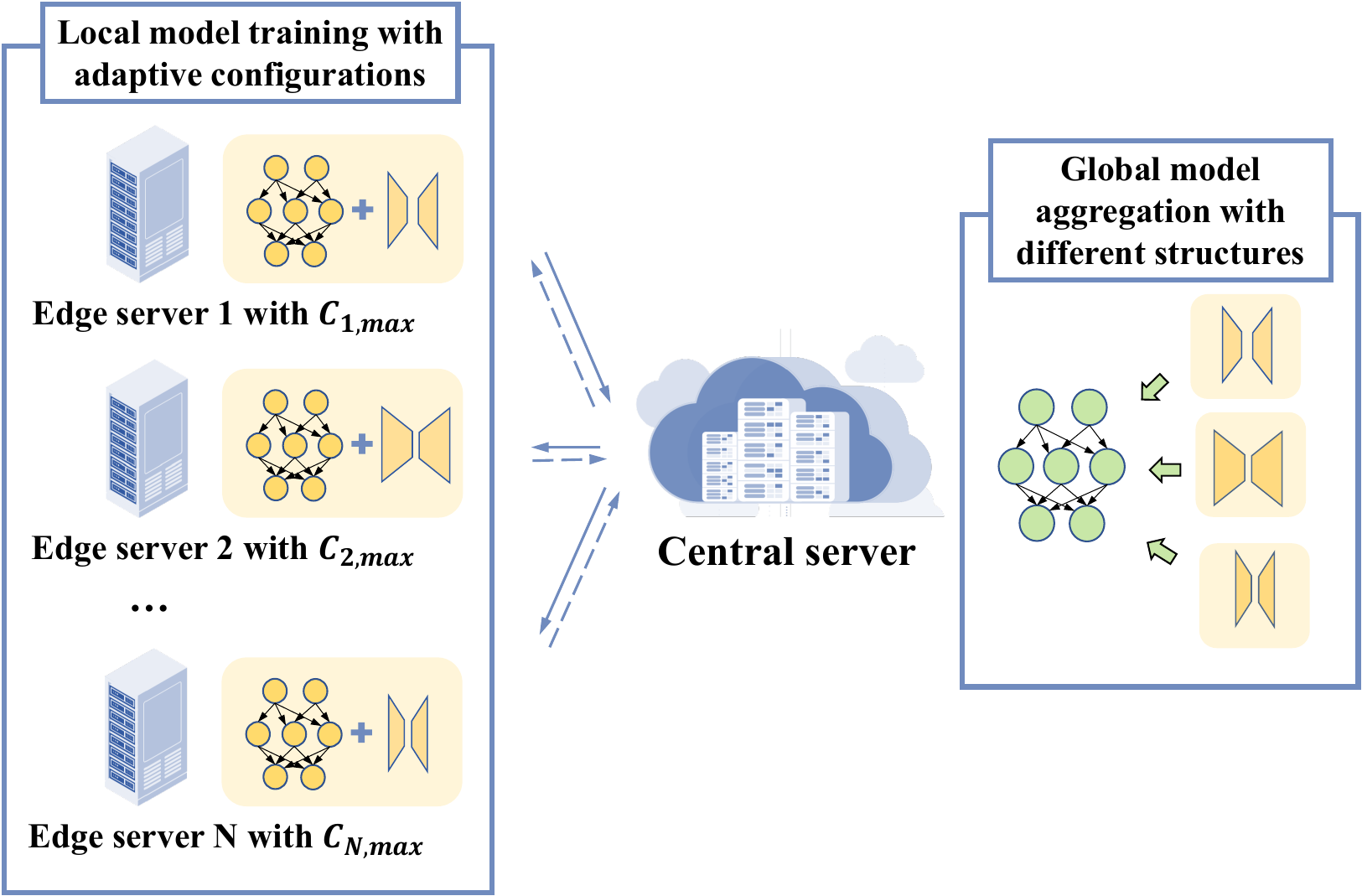}
\caption{Integrating LoRA adapters into LLM FL.}
\label{Scenario}
\vspace{-2ex}
\end{figure}

\newrev{
In this section, we model the LoRA adapters for different edge servers with heterogeneous computing resources. \newrev{As illustrated in Fig.~\ref{Scenario}, $N$ edge servers independently train LoRA adapters, each with its computing budget $C_{i,\max}$ for the $i$-th edge server. After training, the central server aggregates all LoRA adapters of edge servers.}
LoRA is built on the insights that over-parametrized models essentially reside on a low intrinsic dimension~\cite{aghajanyan2021intrinsic}, and thus, can be characterized with the reduced number of trainable parameters.}
% In view of this, we employ the most recent Low-Rank Adaptation~\cite{hu2021lora}, which has advantages in the reduced trainable parameters, the increased training throughput, and no additional inference latency. Building on the insights provided by~\cite{aghajanyan2021intrinsic} that the learned over-parametrized models essentially reside on a low intrinsic dimension, 
LoRA augments the parameters of the frozen \newrev{pre-trained} model with an additional factorized projection, which constrains weight updates by representing them with a low-rank decomposition $\mathbf{W}=\mathbf{W}_0 + \Delta \mathbf{W}$, where \newrev{$ \Delta \mathbf{W} = \mathbf{BA} $, $\mathbf{B} \in \mathbf{R}^{d_i \times r}$ and  $\mathbf{A}\in \mathbf{R}^{r \times d_o}$ denote the incremental matrix and rank decomposition matrices, respectively. The rank is $r\ll \min{(d_i, d_o)}$ (e.g., $r = 8$ when $d_i = d_o = 1024$), and the dimensions of input and output of rank decomposition matrices $\bf B$ and $\bf A$ are $d_i$ and $d_o$, respectively. The LoRA adapter is the cascade structure composed of $\mathbf{B}$ and $\mathbf{A}$.
}

\newrev{
The trainable weight of the $i$-th edge server is denoted by  $\mathbf{W}_i$, and its local training dataset $ \mathcal{D}_{i} = \left \{\mathbf{x}_{i,j}, y_{i,j} \right \}$, where $ \mathbf{x}_{i,j} $ and $ y_{i,j} $ represent the $j$-th input data and its corresponding label, respectively. 
}
% The LoRA adapter is denoted by its weights and bias $\mathbf{W}$, and $i$-th edge server performs local training based on its local dataset $ D_{i} = \left \{ \mathbf{x}_{i,j}, y_{i,j} \right \}$, where $ \mathbf{x}_{i,j} $ and $ y_{i,j} $ represent the $j$th input data and its corresponding label.
% computation based on its local dataset $ D_{i} = \left \{ \mathbf{x}_{i,j}, y_{i,j} \right \}$, where $ \mathbf{x}_{i,j} $ and $ y_{i,j} $ represent the $j$th input data and its corresponding label. 
Thus, the local loss function for \newrev{the $i$-th edge server} can be denoted as $L_{i} (\mathbf{W}_i) = \frac{1}{\left | \mathcal{D}_{i} \right | }  \sum_{j=1}^{\left | \mathcal{D}_{i} \right |} L_{i,j} (\mathbf{x}_{i,j}, y_{i,j}; \mathbf{W}_i)$ where $L_{i,j} (\mathbf{x}_{i,j}, y_{i,j}; \mathbf{W}_i)$ represents the sample-wise loss function for the $j$-th data sample in the local dataset $\mathcal{D}_{i} $. 
Denote the total dataset as $ \mathcal{D} = {\textstyle \bigcup_{i=1}^{N} \mathcal{D}_{i}}$, the global loss function is expressed as the weighted average of the local loss functions. To find the optimal model parameters \newrev{$ \mathbf{W}^{\ast}$,} the learning objective of the global model can be formulated as
\begin{equation} \label{eq:fl_formu}
    \min_{\mathbf{W}} {L(\mathbf{W})} = \min_{\mathbf{W}}\sum_{i=1}^{N}\frac{\left|\mathcal{D}_{i} \right|}{\left| \mathcal{D} \right | }{L_{i}(\mathbf{W})}
\end{equation}
\newrev{
where ${\bf{W}} = \sum\limits_{i = 1}^N {\frac{{\left| {{{\mathcal D}_i}} \right|}}{{\left| {\mathcal D} \right|}}{{\bf{W}}_i}} $.

Recalling Section~\ref{ssec:bg:weight}, batch size and rank affect the LoRA training performance directly, and thus, we rewrite Eqn.~\eqref{eq:fl_formu} in the following way
\begin{subequations}\label{optimization problem}
\begin{alignat}{2}
\min_{b_{i},\mathcal{R}_{i}, \mathcal{W}_{i}} \quad & \sum_{i=1}^{N}\frac{\left|\mathcal{D}_{i} \right|}{\left| \mathcal{D} \right | }{L_{i}(\mathbf{W} | b_{i},\mathcal{R}_{i}, \mathcal{W}_{i})}, & \tag{2a} \label{eq:opt:2a} \\
\mbox{s.t.}\quad
& b_{min} \le b_{i} \le b_{max},  & \tag{2b} \label{eq:opt:2b} \\
& \mathcal{W}_{i} \in \{\mathbf{W}_q,\mathbf{W}_k,\mathbf{W}_v,\mathbf{W}_o \}, &\tag{2c} \label{eq:opt:2c} \\
& {r}_{i}^m  \in  \mathcal{Q}^m, &\tag{2d} \label{eq:opt:2d} \\
& {b_{i} \cdot f_{c}(\mathcal{W}_{i}, \mathcal{R}_{i})\le C_{i,\max}}, &\tag{2e}\label{eq:opt:2e}
\end{alignat}
\end{subequations}
}
where $\mathcal{W}_{i}$ and $\mathcal{R}_{i}=\left \{{r}_{i}^1, {r}_{i}^2,..., {r}_{i}^M \right\}$
represent the sets of selected trainable weights and corresponding ranks, constraint Eqn.~\eqref{eq:opt:2b} refers to the range of the batch size for model training, and $b_i$ is the batch size of the $i$-th edge server. \newrev{The constraint Eqn.~\eqref{eq:opt:2c} is used to pick up important weights in the transformer layer~(see Section~\ref{ssec:bg:weight}).}
\newrev{Eqn.~\eqref{eq:opt:2d} guarantees that the rank of the $m$-th weight is selected in $\mathcal{Q}^m$ where $m$ ranges from 1 to $M=4$ to represent different weights (i.e., $\mathbf{W}_q,\mathbf{W}_k,\mathbf{W}_v,\mathbf{W}_o$). Eqn.~\eqref{eq:opt:2e} indicates the computing budget for the $i$-th edge server where \rev{$f_c(\cdot) $ maps the relationship between trainable weight structure to floating point operations per second (FLOPS)~\cite{kaplan2020scaling}.}} 
% the rank of model parameters is selected in $\mathcal{D}$ set, which considers both model performance and available resource of each edge server.} 
% The constraints in Eqn.~\eqref{eq:opt:2d} and Eqn.~\eqref{eq:opt:2d} indicate that the resource used for training cannot exceed the maximum resources allocated to FL model training for computing and storage respectively, ensuring that every edge server is capable of training these models.
The optimization problem (\ref{optimization problem}) is an MILP problem, which is typically NP-hard. Hence, we propose an automated federated pipeline framework to solve this problem from Section~\ref{ssec:import_identify} to~\ref{ssec:quant}.

\subsection{Identification of important weights} \label{ssec:import_identify}

\newrev{
Recalling in Fig.~\ref{fig:diff_config} (see Section~\ref{ssec:bg:weight}), when the whole training process is done, different weights have various average performances on the LoRA adapter. In practice, the situation is more complex, since different combinations of weights also result in various training performances. More importantly, under a computing budget, the $i$-th edge server needs to identify and predict the combinations of weights with LoRA adapters for the next round of training. 

% \name needs to automatically determine the set of important weights, and train them with LoRA for the $i$-th edge server. 
 
}
\newrev{
To identify a set of important weights for each edge server, \name introduces an indicator for important weights to quantify the important level of parameters and the sensitivity of parameters in the training loss.}
% To quantified the overall contribution of the adapter $A$ and $B$ to model performance, we introduce an important layer indicator 
% which quantifies 
% the sensitivity of parameters to the training loss. 
\rev{The well-known principal components analysis (PCA)~\cite{labrin2020principal} algorithm identifies the most information-rich dimensions with larger singular values by employing matrix singular value decomposition (SVD). Motivated by this, 
% we first utilize the singular values to quantify the importance of the weights, and thus 
we perform the SVD on $\mathbf{BA}$ (see Section ~\ref{ssec:lora}) of each trainable weight before each round of the training. 

Since larger singular values contain richer data information, we characterize the importance of weights using the average singular value of each trainable weight. Additionally, we also devise a sensitivity-based importance scoring function based on sensitivity-based importance scoring~\cite{molchanov2019importance, zhang2022platon, liang2021super}.
} Therefore, at the $t$-th training round, to properly measure the parameter \rev{contribution of the $m$-th trainable weight (LoRA adapter) for $i$-th edge server}, the importance metric can be calculated as follows
\begin{equation}
{s^{m{\kern 1pt},t}_{i}} = \frac{1}{{{d_1}}}\sum\limits_{j = 1}^{{d_1}} {{\lambda^{m{\kern 1pt},t}_{i,j}}}  + {\phi}\left( {{{\bf{W}}^{m{\kern 1pt},t}_{i}}} \right)
\end{equation}
\rev{where $\lambda^{m{\kern 1pt},t}_{i,j}$ denotes the $j$-th singular value of the $m$-th trainable weight (LoRA adapter) for $i$-th edge server at the $t$-th training round, ${{\bf{W}}^{m{\kern 1pt},t}_{i}}$ represents the $m$-th trainable weight (LoRA adapter) for the $i$-th edge server at the $t$-th training round and $d_1$ is the total number of singular values. Here, ${\phi}\left( {{{\cdot}}} \right)$ is a sensitivity-based importance scoring function for model weights. To interpret ${\phi}\left( {{{\cdot}}} \right)$ function, we first define the magnitude of the gradient-weight product:}
% \needrev{where $s_{k,i}$ is the important indicator for the $m$-th LoRA adapter corresponding to the $m$-th weight, and $\mathbf{\Lambda}$, $P$ and $Q$ are the diagonal matrix containing singular values, and left/right singular vectors after undertaking SVD on $BA$, which is expressed as $BA=P \Lambda Q$. }
\begin{equation}\label{eqn:average_gra_wei}
I\left( {\bf{W}} \right) = \frac{1}{d_2}\sum\limits_{j = 1}^{{d_2}} {\left| {{w_j}{\nabla _{{w_j}}}L\left( {\bf{W}} \right)} \right|} 
\end{equation}
\rev{where ${w_j}$ and ${{\nabla _{{w_j}}}L\left( {\bf{W}} \right)}$ represent the $j$-th trainable parameter of the model parameter ${\bf{W}}$ and its corresponding gradient, respectively. $d_2$ is the total number of model parameter $\bf W$. 

Eqn.~\eqref{eqn:average_gra_wei} essentially approximates the average change in loss per parameter when a parameter
is zeroed out~\cite{molchanov2019importance,liang2021super}.  However, the sensitivity in Eqn.~\eqref{eqn:average_gra_wei} is not a reliable indicator of the importance of ${\bf W}$. This is because it is estimated based on one mini-batch data sample, where random sampling and complex training dynamics result in high variability and significant uncertainty in sensitivity estimation using Eqn.~\eqref{eqn:average_gra_wei}.  Therefore, at the $t$-th training round, we address this problem by designing sensitivity smoothing and uncertainty quantification~\cite{zhang2022platon}: }
\begin{equation}
{{\overline I}^{t}}\left( {\bf{W}} \right) = {\lambda _1}{{\overline I}^{t-1}}\left( {\bf{W}} \right) + \left( {1 - {\lambda _1}} \right){I}^{t}\left( {\bf{W}} \right)
\end{equation}
\begin{equation}
{{\overline U}^{t}}\left( {\bf{W}} \right) = {\lambda _2}{{\overline U}^{t-1}}\left( {\bf{W}} \right) + \left( {1 - {\lambda _2}} \right)\left| {{I}^{t}\left( {\bf{W}} \right) - {{\overline I}^{t}}\left( {\bf{W}} \right)} \right|
\end{equation}
\rev{where $0 < {\lambda _1}, {\lambda _2} < 1$, ${\overline I}\left( {{{\cdot}}} \right)$ represents the smoothed sensitivity by exponential moving average, and ${\overline U}\left( {{{\cdot}}} \right)$ denotes the uncertainty term. Subsequently, the sensitivity-based importance scoring function ${\phi}\left( {{{\cdot}}} \right)$ is defined as the product of ${\overline I}\left( {{{\cdot}}} \right)$  and ${\overline U}\left( {{{\cdot}}} \right)$~\cite{zhang2022platon}:}
\begin{equation}
{\phi}^{t}\left( {\bf{W}} \right) = {{\overline I}^{t}}\left( {\bf{W}} \right){{\overline U}^{t}}\left( {\bf{W}} \right)
\end{equation}

\begin{figure}[t]
  \centering
  \subfloat[Normalized important metric.\label{fig:important_indicator}]{
    \includegraphics[width=0.5\linewidth]{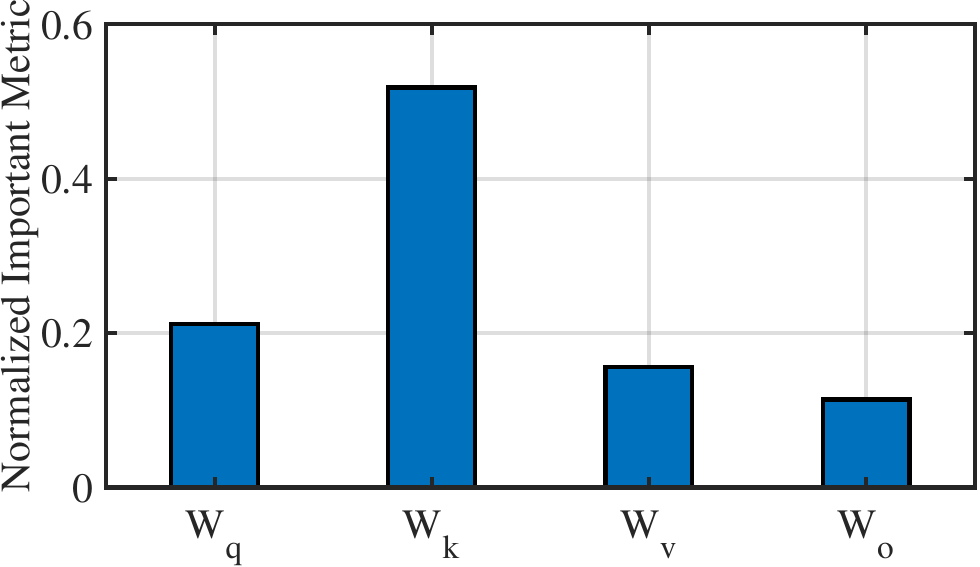}
  }
  \subfloat[PPL\label{fig:important_ppl}]
  {
    \includegraphics[width=0.5\linewidth]{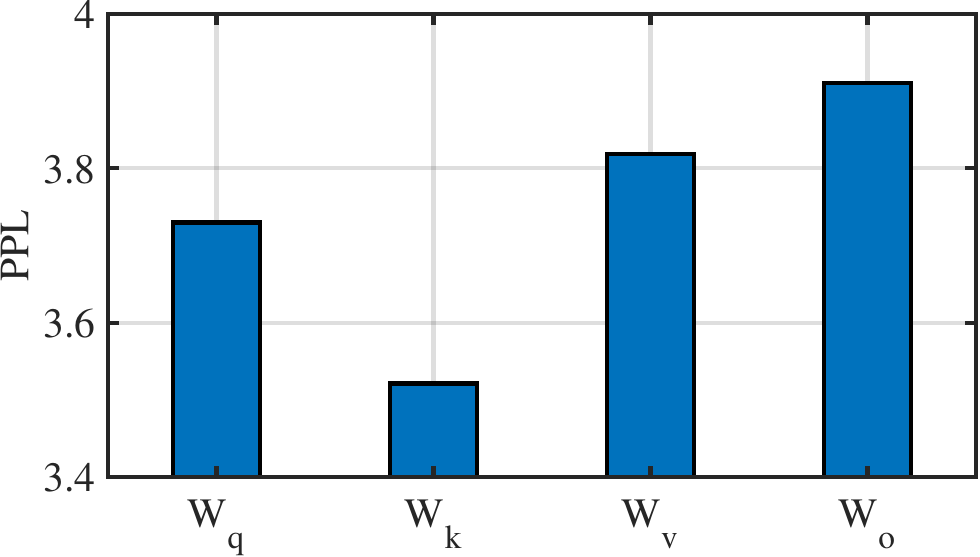}
  }
  \caption{The average normalized important metric and PPL of different trainable weights for GPT-2 models across 20 training rounds.}
  \label{fig:important_metric}
  \vspace{-2ex}
\end{figure}
% \needrev{We calculate important indicators of  study our the important indicator }
\rev{
As shown in Fig.~\ref{fig:important_metric}, we observe that trainable weights with higher importance metrics exhibit better fine-tuning performance, i.e., lower PPL values, which further illustrates the effectiveness of the proposed importance metrics.} \newrev{
With the importance of weights $\mathbf{s}_i = \{s^{m}_{i}$\}, the $i$-th edge server can configure its set of LoRA adapters in the following section.}

% the optimal model performance could be achieved based on the heterogeneous adapters configuration in the following Section.

% \begin{table}
% \centering
% \begin{tabular}{ |c||c|c|c|c|c| }
%  \hline
%  \multirow{2}{*}{Model} &\multicolumn{5}{c|}{Batch Size} \\ &4 &8 &16 &32 &64 \\
%  \hline
%  & 105.62M & 105.62M & 44.0  & 23.4  & 13.1 \\
%  \hline
%  \hline
%  \multirow{2}{*}{Model} &\multicolumn{5}{c|}{Rank} \\ &4 &8 &12 &16 &20 \\
%  \hline
%  & 105.62M & 105.62M & 44.0  & 23.4  & 13.1 \\
%  \hline
% \end{tabular}
% \caption{Trainable Parameters in Bert and GPT-2 model with different training batch sizes and ranks.}
% \label{tab:bert_performance}
% \end{table}

\begin{figure}[t]
    \centering
    \includegraphics[width=0.75\linewidth]{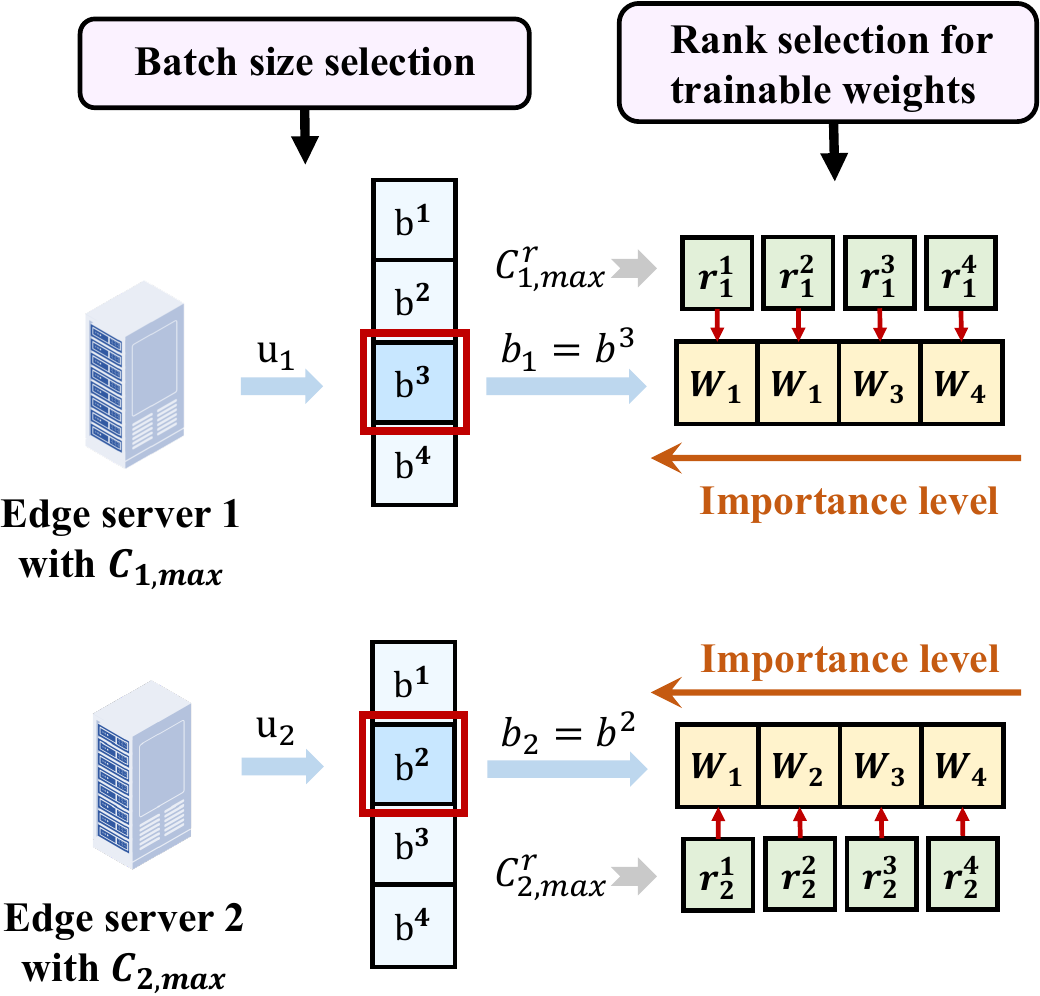}
    \caption{The automated LoRA adapters configuration solution.}
    \label{Optimal_Solution_fed}
    \vspace{-2ex}
\end{figure}

\subsection{Heterogeneous LoRA Adapters Configuration} \label{ssec:adp_cfg}

\newrev{
We propose an automatic configuration solution to select batch size and rank for each edge server, based on the proposed important weight indicators. Since exhaustive search for all kinds of batch sizes, weights, and ranks is impossible due to the large search space, \name leverages a two-stage search algorithm to tackle this problem as shown in Fig.~\ref{Optimal_Solution_fed}. The first stage is to determine batch size, according to data diversity, and the second stage is to find the suitable rank under a computing budget.}    

\needrev{To synchronize model aggregation across various edge servers (aligning the completion time for local model updates from different edge servers), the batch size selection is related to the computing capability at each edge server. For example, two edge servers with the computing capabilities of 20~\!TFLOPs and 40~\!TFLOPs, respectively, have different contributions to the LoRA adapters.} 
% \needrev{We study data diversity impact on batch sizes shown in Figure~\ref{fig:batch_and_rank} that larger data size leading to better contribution to LoRA adapters.} 
\newrev{Consequently, considering $\mathbf{C} = \left[C_{1,max}, C_{2,max}, \cdots, C_{N,max} \right]$ computing capabilities of $N$ edge servers, \name calculates ratios of all values in $\mathbf{C}$, \needrev{expressed as  $ \mathbf{U}_r = [u_1:u_2: \cdots: u_N ]$, and the batch sizes of edge servers are determined by $\mathbf{B} =  \mathbf{U}_r \times b_{\max} $.} It is noted that any value $b_i < b_\min $ in $\mathbf{B}$ is replaced by $b_\min$. }

\newrev{When the batch sizes $\mathbf{B}$ is determined, for the $i$-th edge server, \name can obtain its computing resource constraint only related to rank via $C_{i,\max}^r = C_{i,\max}/b_i$ based on Eqn.~\eqref{eq:opt:2e}. To pick up the ranks and weights from the sorted weights according to the importance~(see Section~\ref{ssec:import_identify}), \name first starts from the most important weight, and then tries the rank value sequentially in $\mathcal{Q}^m$ from high to low. If the computing capability corresponding to the highest ranked weight is below the maximum computing constraint $C_{i,\max}^r$, it is selected. If not, the second-highest-ranked weight is used to evaluate. The above steps repeat for the other weights until sets of weights and ranks reach the computing constraint $C_{i,\max}^r$. At the end, the rank adjustment is iterated for each weight, and the optimal LoRA adapters at each edge server are configured dynamically during every training round.
We summarize the whole process steps in {Algorithm~\ref{alg:fed}}. 
}

\RestyleAlgo{ruled}
\LinesNumbered
\begin{algorithm}
\caption{\name training.}\label{alg:fed}
\setstretch{1.0}
\small
\SetKwInOut{Input}{Require}
\SetKwProg{Fns}{Server Executes}{:}{}
\SetKwFunction{Fns}{Server Executes}
\SetKwProg{Fn}{}{:}{}
\SetKwFunction{Fa}{Auto Configuration}
% \SetKwFunction{Fi}{Importance Identification}

\Input{$N$ is the total number of edge servers, $\bf{U}_r$ is the ratio of computing budgets of $N$ edge servers.}
\KwData{$\{\mathcal{D}_{1}, \cdots, \mathcal{D}_{i}, \cdots, \mathcal{D}_{N}\}$ where $\mathcal{D}_{i}$ is the local collected data on the $i$-th edge servers.}

\Fn{\Fns}{
    initialize the global model $\mathbf{W}^{(0)}_g$ at $t=0$\;
    $S \leftarrow \{C_1, \cdots, C_N \}$\;
    $\mathbf{B} \leftarrow \max(b_{min}, \mathbf{U}_r \times b_{\max})$\;
    \For{communication round $t$}{
    \For{edge server $C_i \in S$ in parallel }{
    % $\mathbf{W}^{t}_i \leftarrow \mathbf{W}^{t}_g$\;
    $\mathbf{B}_{i},\mathbf{A}_{i} \leftarrow \FuncSty{Auto Configuration}(i,b_{i})$\;
    $\mathbf{W}^{t+1}_i \leftarrow \mathbf{W}^{t}_g + \mathbf{B}_{i} \mathbf{A}_{i}$\;
    $\mathbf{W}^{t+1}_i \leftarrow Adam(\mathbf{W}^{t+1}_i, b_{i})$\;
    }
    $\mathbf{W}^{t+1}_g \leftarrow \FuncSty{Model Aggregate}(\mathbf{W}^{t+1}_i)$
    }
}

\Fn{\Fa{$i, b_{i}$}}{
    $C_{i,\max}^r \leftarrow C_{i,\max}/b_i$ \;
    \For{ $m=1:M$ }{
    $s^{m}_{i} \leftarrow \FuncSty{Weight Importance}(\mathbf{B}^{m}_{i}, \mathbf{A}^{m}_{i})$ \;
    }
    $index \leftarrow \FuncSty{Sort_decent}(\mathbf{s}_i)$ \;
    \For{ $m=index(1):index(M)$ }{
    \For{ $r=\mathcal{Q}^m_{max}:\mathcal{Q}^m_{min}$ }{
    $\mathcal{W}^{m}_{i} \leftarrow \mathbf{W}^{m{\kern 1pt}}_i + \mathbf{B}^{m}_{i} \mathbf{A}^{m}_{i}$ \;
    \uIf{$f_{c}(\mathcal{W}^{m}_{i},r)\le C_{i,\max}^r$}{
    % ${r}_{i}^{m} \leftarrow r$\;
    $\mathbf{B}^{m}_{i},\mathbf{A}^{m}_{i} \leftarrow$ set LoRA rank to $r$ \;
    $C_{i,\max}^r \leftarrow C_{i,\max}^r - f_{c}(\mathcal{W}^{m}_{i},r)$
    }
    \uElse{
    continue\;
    }
    }
    }
    \KwRet $\mathbf{B}_{i}, \mathbf{A}_{i}$\;
}
\end{algorithm}

\subsection{Quantization for Adapters' Training} \label{ssec:quant}
\rev{
Apart from the LoRA adapter configuration, the limited and heterogeneous GPU memory resources of edge servers may also lead to an increase in the under-training rate, thus posing a significant challenge for LLM fine-tuning. Although model quantization has emerged as a promising solution to address memory constraints, existing methods do not consider varying quantization bits across edge servers.

%Although model quantization has emerged as a promising solution to address memory constraints, existing studies primarily focus on compressing model for inference, neglecting the potential effect of quantization in the training process. 

Considering the heterogeneous GPU memory \newrev{budgets} of edge servers ($M_{i, max}$, $i=1,2,...,N$), we quantize the pre-trained models with varying quantization bits, and only de-quantize models for performing matrix multiplication with higher precision. \newrev{
Pre-trained models consume a significant amount of memory space, necessitating a high compression ratio to conserve memory (usually 4 bits or 8 bits). In contrast, LoRA adapter requires less memory space and is utilized for updates, hence a lower compression ratio (e.g., 16-bit floats (FP16)) is employed to maintain precision. Therefore, in \name, we quantize pre-trained models with the maximum quantization bits based on memory budgets of edge servers, but keep LoRA adapters unchanged. In the following, we give details of our quantization.
}

% Since most \newrev{of memory footprint} for LLM fine-tuning stems from the pre-trained model rather than LoRA adapter parameters, we quantize the pre-trained model with low-precision (usually 4 bits or 8 bits) and remain LoRA adapter as 16FP to achieve better performance while reducing the memory footprint.

}

\rev{
% Quantization is the process of converting data type from the higher-information representation with high-bit to representation with low-bit, such as compressing 32-bit floats (FP32) to 8-bit integers. 
\newrev{In our design,} we adopt the NormalFloat (NF) quantization built based on Quantile Quantization~\cite{dettmers20218}, which offers significant benefits in model compression, computational complexity reduction, accuracy preservation, and hardware compatibility~\cite{dettmers2023qlora}. NF data type ensures that each quantization bin has an equal number of values for the input tensor. It estimates the quantile through the empirical cumulative distribution function, enhancing quantization accuracy. 

% Considering 
\newrev{Since pre-trained model} weights usually follow a zero-mean normal distribution~\cite{dettmers2023qlora}, we first transform all weights to a zero-mean normal distribution within $[-1, 1]$ range via standard deviation. After that, we estimate the $2^k + 1$ quantiles
to obtain a $k$-bit quantile quantization data type and normalize the value of this data type into the $[-1, 1]$ range. Finally, we quantize an input weight tensor by normalizing it into $[-1, 1]$ range through absolute maximum rescaling. Once the weight range and data type range match, \needrev{we estimate the $2^k$ values $q_i$ of the data type as follows}
\begin{equation}
{q_i} = \frac{1}{2}\left[{Q\left({\frac{i}{2^k+1}}\right)}+{Q\left({\frac{i}{2^k+1}}\right)}\right]
\end{equation}
where $Q (\cdot)$ is the quantile function of the standard normal distribution. \newrev{By this means, the pre-trained model quantization is done.} 
% Indeed, our algorithm is straightforward: we quantize pre-trained models with the maximum quantization bits that GPU memory budget of edge servers can accommodate.

}

\subsection{LoRA Adapters Aggregation and Deployment} \label{ssec:aggregate}
\begin{figure}[t]
\centering
\includegraphics[width=\linewidth]{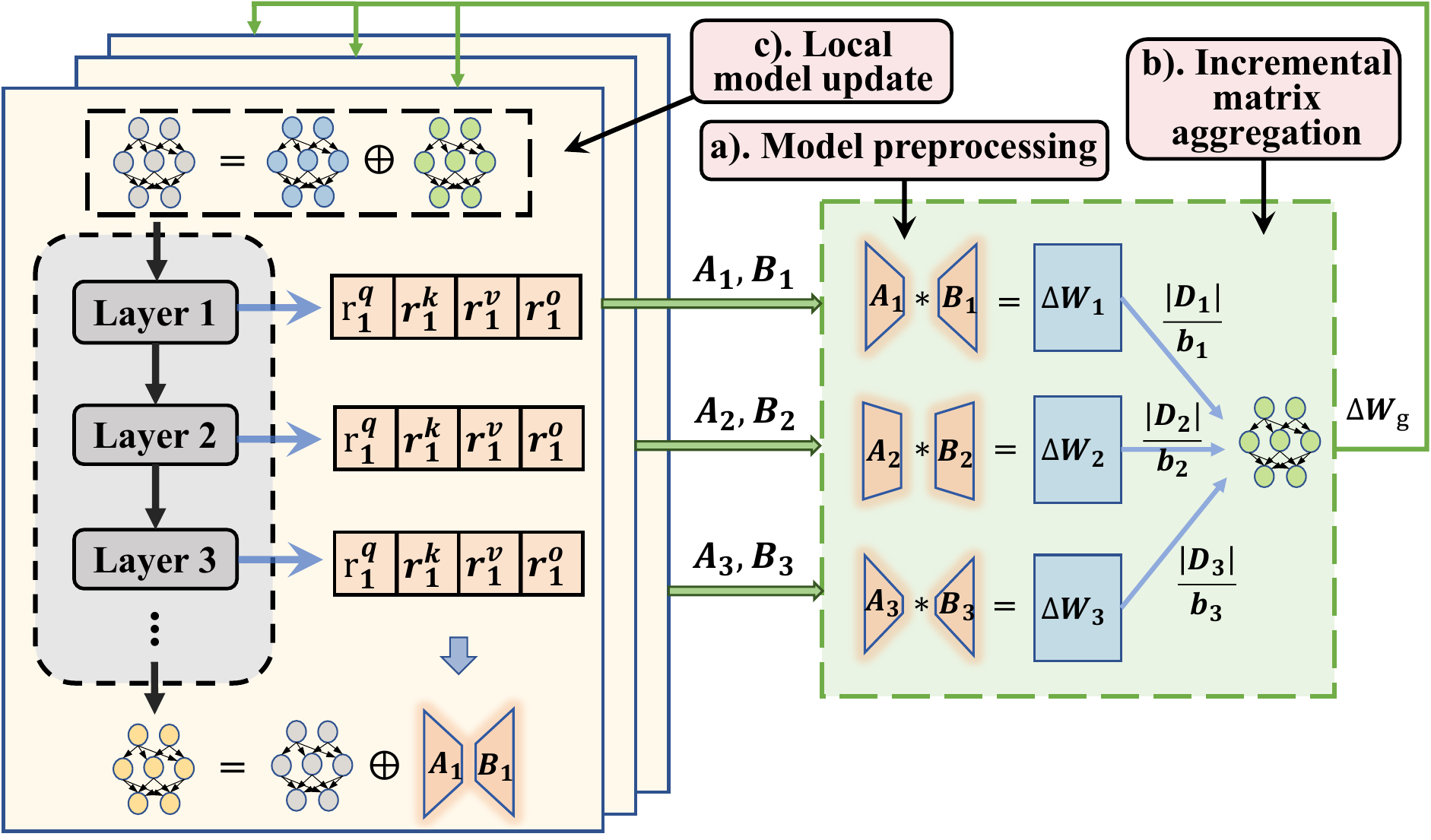}
\caption{Adapter aggregation of \name framework.}
\label{fig:aggregation}
\vspace{-2ex}
\end{figure}

\rev{
% The heterogeneous configuration method in Sec.~\ref{ssec:adp_cfg} configures varying LoRA adapters structures across edge servers. However, traditional FL aggregation methods, such as FedAvg, are only applicable to model aggregation with the same structure, which cannot be directly applied to \name. Therefore, it is imperative to develop an efficient model aggregation method to aggregate heterogeneous LoRA adapter structures from different edge servers.

\newrev{Current FL model aggregation aggregates the whole models, but for LLM, transferring them to the central server brings a significant communication overhead (see Section~\ref{ssec:bg:straggler}). Therefore, \name only allows each edge server to transfer rank decomposition matrices $\mathbf{B}$ and $\mathbf{A}$ (see Section~\ref{ssec:lora}) instead of the whole models in order to reduce the communication overhead. 
However, to merge all kinds of rank decomposition matrices from varying edge servers, we still face two challenges in aggregation and updating, respectively.
% merging all kinds of rank decomposition matrices from varying edge servers is still hard, due to the heterogeneous configuration method in \newrev{Section~\ref{ssec:adp_cfg}}. 
% We face two challenges in aggregation and updating, respectively. 
On the one hand, the diverse rank
decomposition matrices with different ranks from edge servers cannot be aggregated directly~(e.g., different sizes of two edge servers' $\mathbf{B}$s cannot be added). On the other hand, even if the local models are aggregated, the central server recovers $\mathbf{B}$ and $\mathbf{A}$ for the edge server possibly leading to deterioration rather than improvement in the training performance. To combat above challenges,} as illustrated in Fig.~\ref{fig:aggregation}, the tailored heterogeneous model aggregation solution comprises the following three stages.

% We propose a novel model aggregation method to merge heterogenous LoRA adapter structures across edge servers obtained in \newrev{Section~\ref{ssec:adp_cfg}}. The diverse LoRA adapter structures across edge servers pose a significant challenge for directly aggregating rank decomposition matrices. Fortunately, edge servers have consistent dimensions for the incremental matrices of the same trainable weights, providing the basis for the design of our model aggregation scheme. Therefore, instead of directly aggregating rank decomposition matrices, we aggregate the incremental matrices of the same trainable weights from edge servers. This design enables efficient aggregation by transforming heterogeneous LoRA adapters into uniform dimensions. 

% Moreover, since rank decomposition matrices cannot be accurately recovered from the aggregated incremental matrix, directly updating adapters may lead to deterioration rather than improvement in training performance. Thus, we aggregate the incremental matrices into pre-trained models for model updates, avoiding the overhead and performance degradation caused by the inaccurate decomposition of incremental matrices. As illustrated in Figure~\ref{fig:aggregation}, the tailored heterogeneous model aggregation scheme comprises the following three stages.

{\bf a) Model Preprocessing.}  
% After receiving updated rank decomposition matrices $\bf B$ and $\bf A$ from participating edge servers, the central server multiplies these decomposition matrices to obtain their corresponding incremental matrices~\newrev{(see Section~\ref{ssec:lora})}. 
\newrev{Before aggregation, the central server achieves incremental matrices for participating edge servers.}
At the $t$-th training round, the incremental matrix of the $m$-th trainable weight (LoRA adapter) for $i$-th edge server is given by
\begin{equation}
\triangle \mathbf{W}_{i}^{m {\kern 1pt},t} = \mathbf{B}_{i}^{m{\kern 1pt},t} \cdot \mathbf{A}_{i}^{m{\kern 1pt},t}
\end{equation}
where $\mathbf{B}_{i}^{m{\kern 1pt},t}$ and $\mathbf{A}_{i}^{m {\kern 1pt},t}$ denote the corresponding rank decomposition matrices. 
% of the $m$-th trainable weights for $i$-th edge server at the $t$-th training round.

{\bf b) Incremental Matrix Aggregation.} At this stage, the central server aggregates the preprocessed incremental matrices from different edge servers, i.e.,
\begin{equation}
\triangle \mathbf{W}^{m{\kern 1pt},t} = \sum_{i=1}^{N} \frac{\left | D_{i} \right | / b_{i}}{ {\textstyle \sum_{i=1}^{N}} \left | D_{i} \right | / b_{i}} \triangle \mathbf{W}_{i}^{m{\kern 1pt},t}
\end{equation}

}

% b). synchronous aggregation: At this step, a core cloud is used to gather information and aggregate model parameters until all the edge servers involved in the model training complete their local model assembles. After $k$th training round, the latest model aggregation in the core server is updated via

% where $\left | D_{i} \right | / b_{i}$ is calculated as the number of local samples feeds in the $k$th training round. Since the times of loss backward in a local training round is different due to the varying size of local datasets and training batches, the involvement of local weight in the global model also make a difference. Thus, $\frac{\left | D_{i} \right | / b_{i}}{ {\textstyle \sum_{i=1}^{N}} \left | D_{i} \right | / b_{i}}$ is considered as the local weight contribution in model aggregation.

% In addition, beneficial from the frozen pre-trained model, only the accumulated gradient update $\triangle \mathbf{W}_{l,i}^{k}$ and the number of local samples are demanded in model aggregation phase, avoiding transmission of all model parameters for global model construction, hence significantly shorten the transmission as well as the communication overhead.
\rev{
{\bf c) Local Model Update.} 
% Since rank decomposition matrices ${\mathbf A}_{i}^{m {\kern 1pt},t}$ and ${\mathbf B}_{i}^{m {\kern 1pt},t}$ cannot be accurately recovered from the aggregated incremental matrix $\triangle \mathbf{W}_{i}^{m {\kern 1pt},t}$, directly update adapters as usual may lead to deterioration rather than improvement in training performance. 
\newrev{
Instead of updating $\mathbf{B}_{i}^{m {\kern 1pt},t}$ and $\mathbf{A}_{i}^{m {\kern 1pt},t}$ directly, the $i$-th edge server updates $\triangle \mathbf{W}_{i}^{m {\kern 1pt},t}$ based on $\triangle \mathbf{W}^{m{\kern 1pt},t}$ dispatched by the central server. Moreover, recalling Section~\ref{ssec:quant}, according to the memory constraint of the $i$-th edge server, the local model is quantized, and hence, $\triangle \mathbf{W}^{m {\kern 1pt},t}$ should be quantized for merging it into the local model. The updating policy is followed by
}
% The incremental matrix $\triangle \mathbf{W}_{i}^{m {\kern 1pt},t}$ can be regarded as the model update of the pre-trained model parallel to the Lora adapter. Therefore, we absorb the updated incremental matrix $\triangle \mathbf{W}_{i}^{m {\kern 1pt},t}$ into the pre-trained model $\mathbf{W}_{i}^{m {\kern 1pt},t}$ and reinitialize the LoRA adapters $\mathbf{B}_{i}^{m {\kern 1pt},t}$ and $\mathbf{A}_{i}^{m {\kern 1pt},t}$ for the $(t+1)$-th training round, which is expressed as 
\begin{equation}
{\mathbf{W}_{i}^{m{\kern 1pt},t+1}} = \mathbf{W}_{i}^{m{\kern 1pt},t} + quant(\triangle \mathbf{W}^{m{\kern 1pt},t}), 
\end{equation}
\begin{equation}
{\mathbf{B}_{i}^{m{\kern 1pt},t+1}} = \mathbf{\overline B}_{i}^{m}, {\mathbf{A}_{i}^{m{\kern 1pt},t+1}} = \mathbf{\overline A}_{i}^{m}
\end{equation}
where \newrev{$\mathbf{\overline A}_{i}^{m}$ and $\mathbf{\overline B}_{i}^{m}$ are the reinitialized rank decomposition matrices, and $quant(\cdot)$ means the quantization function.} 
% Note that different quantization bits are employed to mitigate storage resource constraints for edge servers. Therefore, it is crucial to quantize $\triangle \mathbf{W}_{i}^{m {\kern 1pt},t}$ before merging it into pre-trained model parameter $\mathbf{W}_{i}^{m {\kern 1pt},t}$.  Based on the aforementioned stages, heterogeneous LoRA adapter structures are efficiently aggregated.
}

% When deploying the global model on edge servers after training, the inference can be performed as usual with an explicit addition of $W$ and $BA$, owing to the same shape of both primary weight and the combination of law-rank adapter $B$ and $A$. Meanwhile, the minor memory overhead as well as the latency in inference phase are guaranteed by avoiding extra adapter layers injected in the original model.  

\section{Implementation and Experimental Setup} \label{sec:impl}

% \begin{figure}
%     \centering
%     \includegraphics[width=0.4\linewidth]{Figures/hardware.jpg}
%     \caption{\needrev{The hardware servers for experiments, which contain 16 NVIDIA GeForce RTX 3090 GPU in total and each GPU runs one model training.}}
%     \label{fig:hardware}
% \end{figure}

\quad{\bf{Models:}} We deploy two well-known LLMs, LLaMA2~\cite{touvron2023llama} and GPT-2~\cite{radford2019language}. The LLaMA2 model has been widely used for various natural language understanding tasks, including text generation, summarization, and questioning and answering. The GPT-2 is famous for its ability to generate coherent and relevant content, making it well suited for text generation tasks, such as text completion and language translation. The LLaMA2-7b model has 32 layers and 3.52 billion parameters, whereas the GPT-2 medium and large models have 24 layers and 355 million parameters and 36 layers and 774 million parameters, respectively.

{\bf{Datasets and tasks:}} We evaluate the training performance of \name for instruction-following (IF) tasks on Alpaca dataset~\cite{alpaca} and natural language generation (NLG) tasks on E2E dataset~\cite{novikova2017e2e}. The Alpaca dataset comprises approximately 52,000 samples which spans diverse domains such as philosophy, economics, sociology, and science, while the E2E dataset consists of around 42,000 training, 4,600 validation and 4,600 test examples from the restaurant domain. We fine-tune the LLaMA2-7B model on the Alpaca dataset for IF task, which outputs detailed and contextually relevant responses based on given instructions. Similarly, we train the GPT-2 model on the E2E dataset for NLG task, aiming to transform various forms of inputs into human-readable natural language.

{\bf{Benchmarks:}} We compare \name with three classical benchmarks: (1)~{\bf Vanilla Fine-Tuning (FT)}: FT fine-tunes full parameters of LLMs on each edge server, which is the default fine-tuning technique in most NLP literature~\cite{kenton2019bert}. (2)~{\bf Low-Rank Adaptation (\lora):} \lora introduces low-rank trainable modules ~\cite{hu2021lora} in parallel to weight matrices for LLM fine tuning. Training only low-rank trainable modules while freezing the rest of model parameters substantially reduces the number of trainable parameters without incurring additional inference latency. (3)~{\bf \FedAdp:} \FedAdp~\cite{cai2023efficient} is an edge-cutting FL framework tailored for efficient LLM fine-tuning. It identifies the optimal adapter configuration during the training process, catering to the demands of fine-tuning LLMs in a privacy-preserving manner while minimizing the training costs. To ensure fair comparisons, all benchmarks adopt the same model aggregation algorithm (FedAvg~\cite{mcmahan2017communication}) and random client sampling scheme, consistent with the default settings in prior FL literature~\cite{cai2023efficient, lin2022fednlp}.

% {\bf{Heterogeneity:}} explain the relationship between computing capacity and rank (give a specific mapping function), illustrate the selection range of rank and the distribution of servers' computing capacities.

{\bf{Hardware and hyper-parameters:}} We conduct experiments utilizing 16 NVIDIA GeForce RTX 3090-Ti GPU with the maximum memory of 24\!~GB. 
\needrev{For resource heterogeneity, we constrain the each GPU's maximum memory available for model training by randomly selecting values from a uniform distribution between 12\!~GB and 24\!~GB.}
% \needrev{as shown in Figure~\ref{fig:hardware}.} 
To ensure fair comparisons, we retain the consistent settings for \name and benchmarks across different tasks. For IF task on the Alpaca dataset, mini-batch size, learning rate, and maximum sequence length are set to 2, 0.0001, and 384, respectively. Similarly, for the NLG task on the E2E dataset with GPT-2 medium, we set mini-batch size, learning rate, and maximum sequence length to 8, 0.0002, and 512, respectively. Considering the limited 24\!~GB memory of the NVIDIA GeForce RTX 3090-Ti, we reduce the mini-batch size to 4 while keeping other hyperparameters unchanged when training the GPT-2 large model. The number of selected participating edge servers is set to \needrev{15} and the rank is set to 4 by default unless specified otherwise. All edge servers run in synchronized mode~\cite{ho2013more}.

\section{Evaluation} \label{sec:eval}
This section provides numerical results to evaluate the training performance of \name framework and the effectiveness of each meticulously designed component.

\subsection{Overall Performance Evaluation}

\begin{figure}[t]
  \centering
  \subfloat[LLaMA2 \label{fig:convergence_statics_llama}]{
    \includegraphics[width=0.495\linewidth]{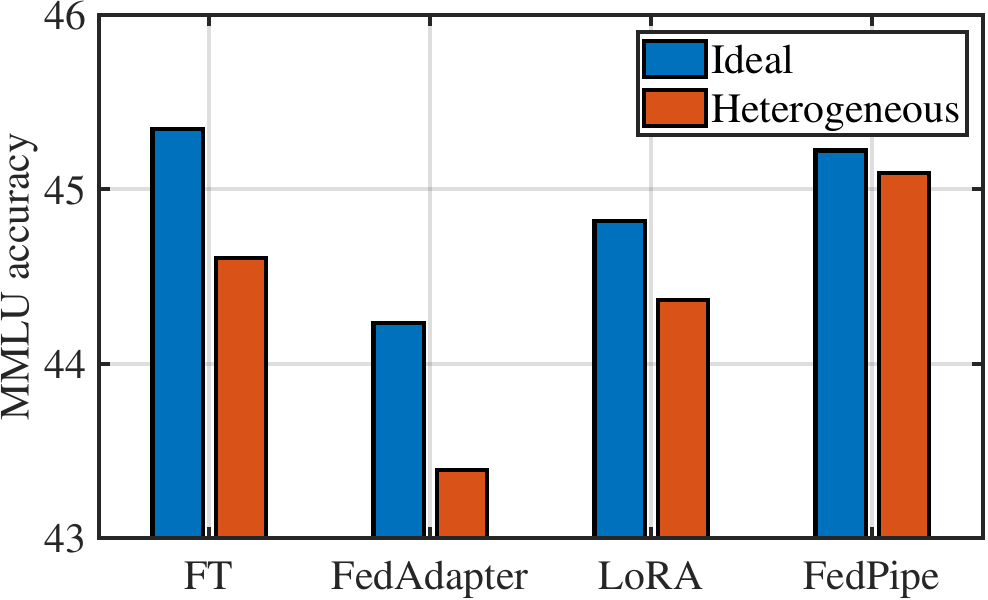}
  }
  \subfloat[GPT-2\label{fig:convergence_statics_gpt2}]
  {
    \includegraphics[width=0.495\linewidth]{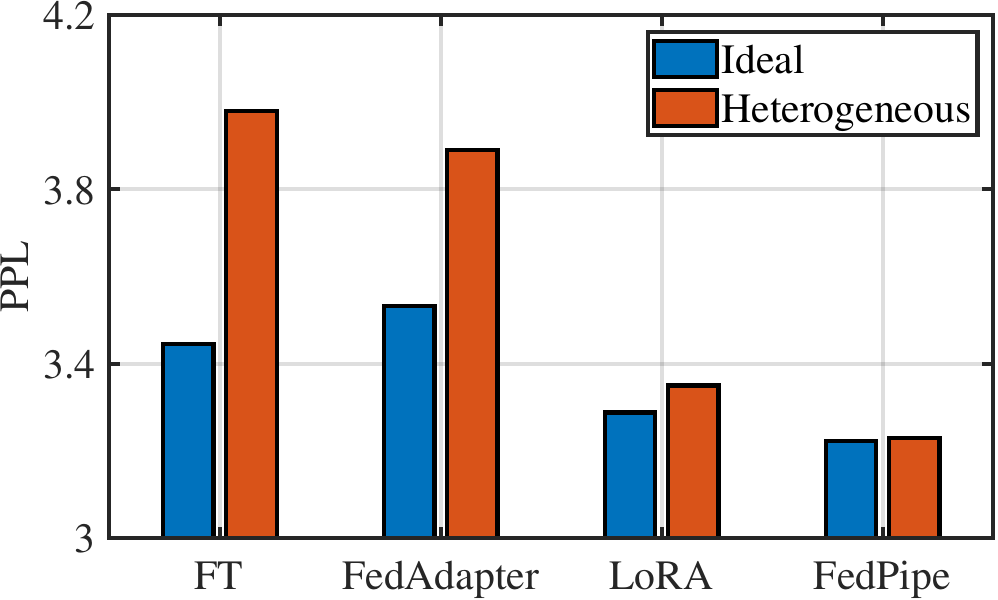}
  }
  \caption{The converged accuracy for LLaMA2-7b and GPT-2 models under heterogeneous and ideal settings.}
  \label{fig:converged_acc}
\end{figure}

\begin{figure}[t]
  \centering
  \subfloat[LLaMA2\label{fig:trainable_parameters_llama}]{
    \includegraphics[width=0.495\linewidth]{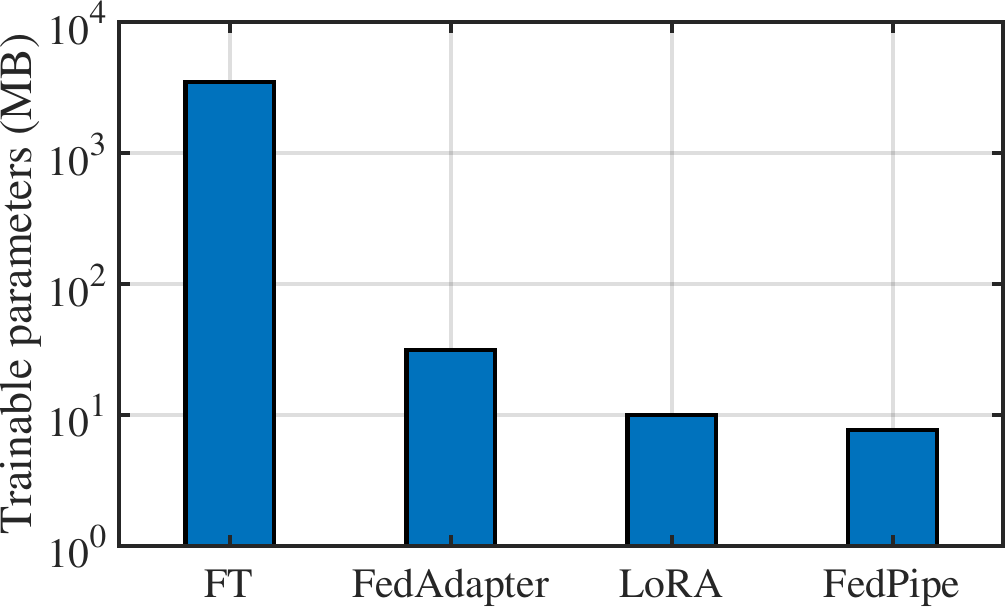}
  }
  \subfloat[GPT-2\label{fig:trainable_parameters_gpt2}]
  {
    \includegraphics[width=0.495\linewidth]{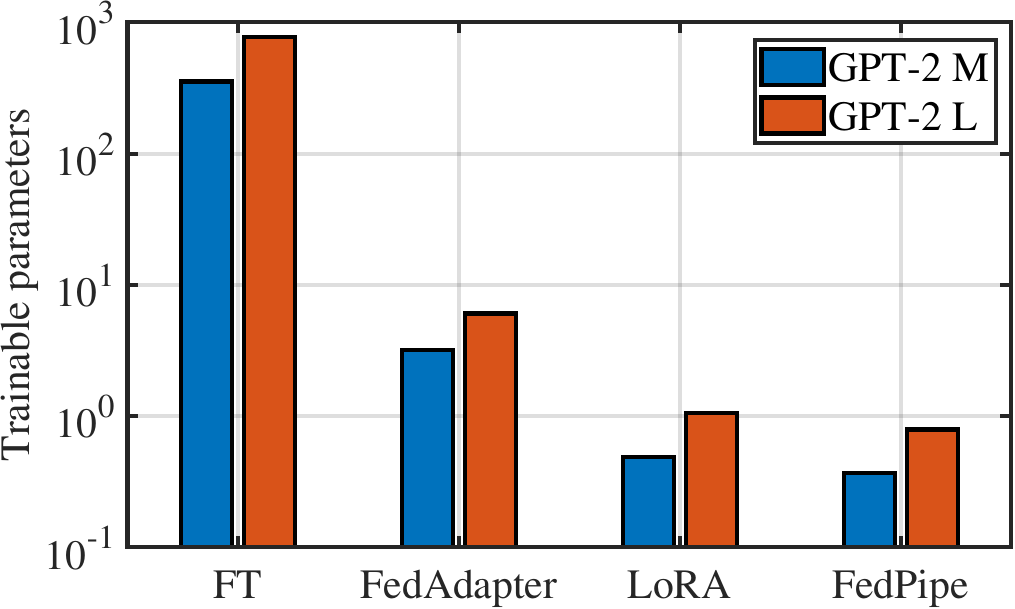}
  }
  \caption{The number of trainable parameters for LLaMA2-7b and GPT-2 models \needrev{under heterogeneous setting}, respectively.}
  \label{fig:trainable_parameters}
  % \vspace{-3ex}
\end{figure}

\begin{figure}[t]
  \centering
  \subfloat[LLaMA2 (ideal). \label{fig:convergence_homo_llama}]
  {
    \includegraphics[width=0.5\linewidth]{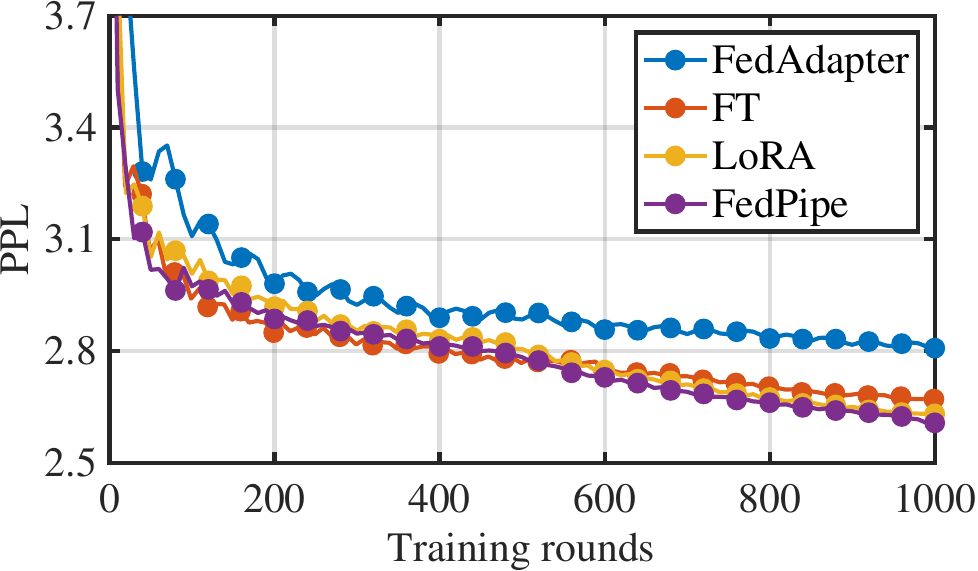}
  }
    \subfloat[LLaMA2 (heterogeneous). \label{fig:convergence_hetero_llama}]
  {
    \includegraphics[width=0.5\linewidth]{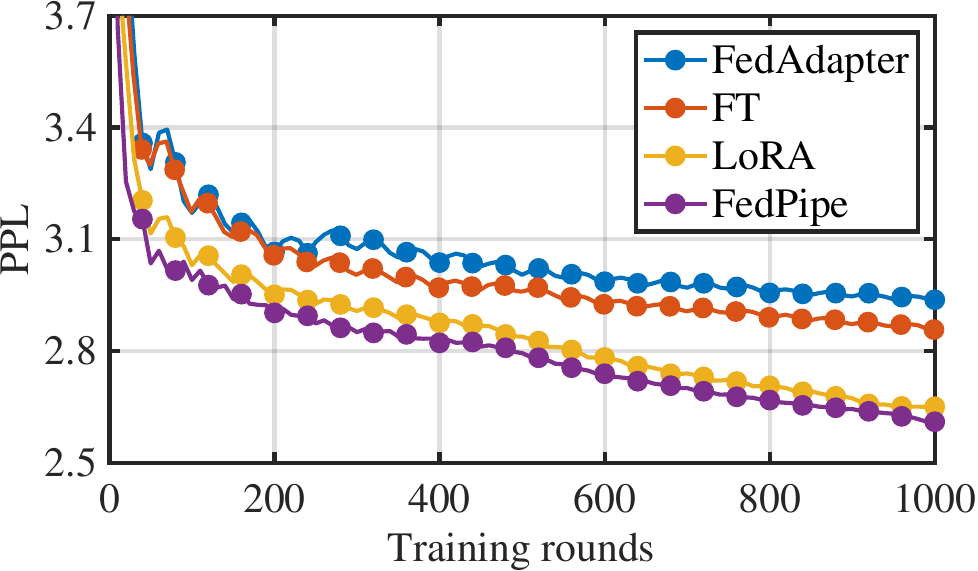}
  } \\
  \subfloat[GPT-2 (ideal).
  \label{fig:convergence_homo_gpt2}]
  {
    \includegraphics[width=0.49\linewidth]{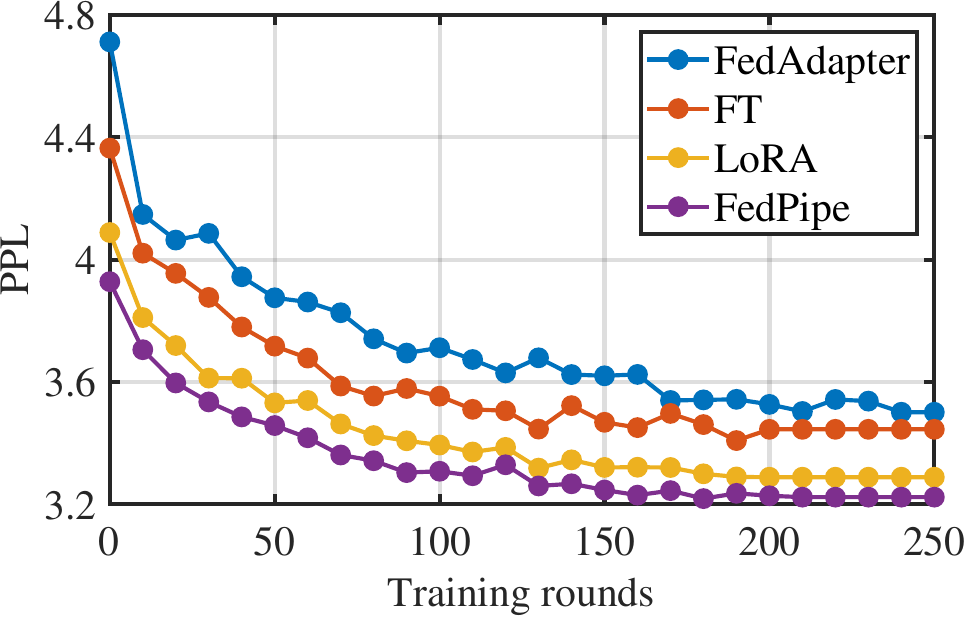}
  }
  \subfloat[GPT-2 (heterogeneous). \label{fig:convergence_hetero_gpt2}]
  {
    \includegraphics[width=0.5\linewidth]{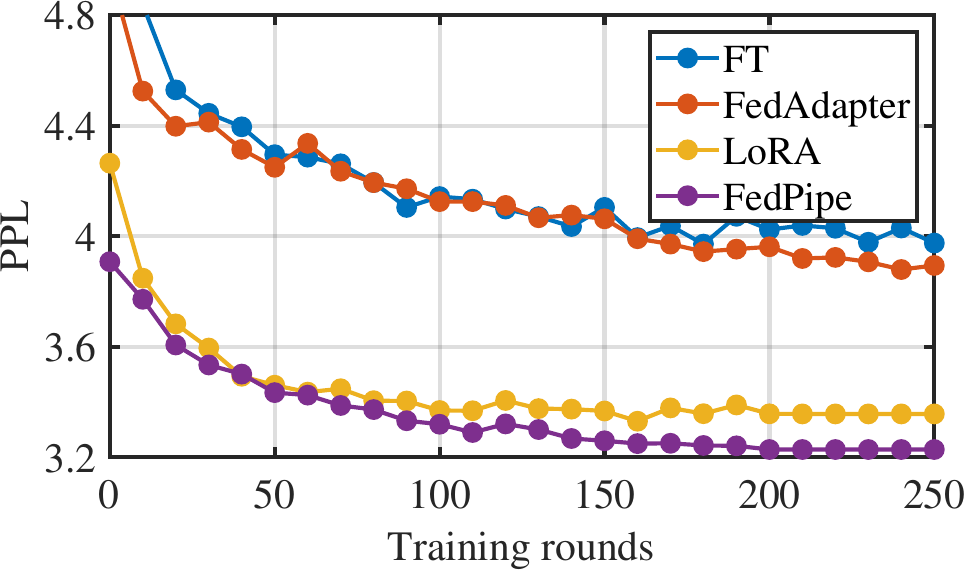}
  }
  \caption{The convergence rate for LLaMA2-7b and GPT-2 models under heterogeneous and ideal settings.}
  \label{fig:converged_rate}
  \vspace{-2ex}
\end{figure}

\quad{\bf{Converged accuracy. }} Fig.~\ref{fig:converged_acc} shows the comparison between \name and the three other benchmarks in converged accuracy on LLaMA2 and GPT-2 models under heterogeneous and ideal settings. It is clear that the proposed \name framework outperforms the other benchmarks in both settings. In the homogeneous setting, the \name framework \rev{reaches comparable accuracy on LLaMA2 model} and outperforms the FT, FedAdapter, and LoRA benchmarks for PPL by \needrev{0.22, 0.31, and 0.06}, respectively. This superiority is attributed to tailored important \rev{weights} identification, which prioritizes the fine-tuning of key \rev{weights}, thereby enhancing the efficiency of fine-tuning LLMs and achieving higher converged accuracy. Moreover, due to the lack of adaptive rank adjustment for computing heterogeneity, all benchmarks exhibit notably inferior converged accuracy under ideal than heterogeneous settings. In contrast, the \name framework automatically adjusts rank size based on heterogeneous computing resources and only experiences a slight increase in PPL, nearly \needrev{0.01}, under heterogeneous than ideal settings. Moreover, the performance gap between \name and the other benchmarks is much larger \rev{than that} in ideal settings, further demonstrating the superior adaptability of \name to heterogeneous scenarios.

% To achieve the same target accuracy as \FedAdp (100\% relative target accuracy for 20NEWS is 0.8 ~\cite{cai2023efficient}) to ensure fair comparisons, we set the rank size of LoRA metrics to 8 and $\alpha$ value (used to adjust the learning rate) to 160 in \name, where all original model parameters are frozen and only the query and value in self-attention layers are inserted with low rank adapters. 

{\bf{Trainable parameters. }}
Fig.~\ref{fig:trainable_parameters} presents the number of trainable parameters for deploying \name and other benchmarks on LLaMA2 and GPT-2 models under heterogeneous and ideal settings. In the ideal setting, we observe that \name framework and LoRA with fixed rank size exhibit identical number of trainable parameters. This is due to \name's capability in fine-tuning all four trainable weights and selecting the maximum rank for each weight without resource constraints. \name has the fewest number of trainable parameters under heterogeneous setting, which is \needrev{nearly 25\%} lower than that under the ideal setting. Deploying \name on LLaMA2 and GPT-2 models results in approximately \needrev{7.69M and 0.37M} trainable parameters, 
% which are about \needrev{1/1200 and 1/30} of that for FT and \FedAdp framework, respectively. 
\rev{whereas the trainable parameters for FT and \FedAdp framework is over 350 and 4 times larger than \name on LLaMA2 model and about 960 and 8 times on GPT-2 models, respectively.}
The reduction in trainable parameters significantly diminishes the computing and communication overheads of fine-tuning LLMs, facilitating the deployment of LLMs. It is noteworthy that \name does not require iterative searching for the optimal depth and width of the adapter layer, reducing the time overhead of the training process. In contrast, \FedAdp selects adapters from three different configuration groups, whereas \name only adapts an initialized adapter configuration, and thus substantially saves storage resources compared to \FedAdp.

{{\bf{Convergence rate. }} 
Fig.~\ref{fig:converged_rate} compares the convergence rate for \name and the other three benchmarks under heterogeneous and ideal settings.  In the ideal and heterogeneous setting, \name exhibits the fastest convergence speed, outpacing FT, FedAdapter, and LoRA by factors of \rev{6, 2.7, and 2 to achieve 3.6 PPL as well as 24, 18 and 2} to achieve 4.0 PPL on GPT-2 model, respectively. Moreover, \name demonstrates comparable performance to FT, and is only
\rev{lower by \needrev{0.23} on LLaMA2 model and even outperforms FT by 0.22 in the metric of PPL on GPT-2 model under ideal setting.}
% \needrev{2.3\%} lower in the metric of PPL. 
This primarily stems from our design of the important layer identification strategy, which prioritizes the fine-tuning of crucial layers to reduce the under-training rate, thus expediting model training. Conversely, in the heterogeneous setting, \name surpasses other benchmarks in convergence rate and accuracy due to its adaptive adjustment of LoRA's rank size to accommodate the heterogeneous computing resources of clients. Due to the absence of heterogeneous LoRA adapter configuration, the convergence speed and accuracy of FT, FedAdapter, and LoRa rapidly deteriorate in heterogeneous settings. For the GPT-2 models, we also conducted a comprehensive comparison of the precision across various metrics under homogeneous and heterogeneous settings, which is summarized in Table~\ref{tab:GPT2_performance}. }

\begin{table*}[t]
\centering
\setlength{\tabcolsep}{12pt}
\scalebox{0.8}{
\begin{tabular}{ |c|c||c||c|c|c|c|c| }
% |p{1.3cm}|p{1.6cm}||p{1.6cm}||p{1.6cm}|p{1.6cm}|p{1.6cm}|p{1.6cm}|p{1.6cm}| }
 \hline
 \multirow{2}{*}{Model} &\multirow{2}{*}{Method} &\multirow{2}{1.6cm}{\#Trainable Parameters} &\multicolumn{5}{c|}{E2E NLG Challenge} \\ & & &BLEU &NIST &MET &ROUGE-L &CIDEr \\
 \hline
 \multirow{5}{1.3cm}{GPT-2 M (Ideal)} 
 &{FT}       & 354.92M & 68.2 & 8.62 & 46.2 & 71.0 & 2.47 \\
 % &\textit{Adapter_\text{L}}   & 11.09M & xxx & 68.9 & 8.71 & 46.1 & 71.3  & 2.47 \\
 &{\FedAdp}  & 3.19M   & 65.4 & 8.26 & 41.9 & 68.6 & 2.06 \\
 &{\lora}    & 0.49M   & 69.1 & 8.74 & 46.3 & 70.6 & 2.46 \\
 &{\name}    & 0.49M   & 69.6 & 8.79 & 46.9 & 70.6 & 2.44 \\
 \hline
 \multirow{5}{1.3cm}{GPT-2 M (Hetero)} 
 &{FT}       & 354.92M & 60.5 & 8.12 & 41.0 & 64.4 & 1.92 \\
 % &\textit{Adapter_\text{L}}   & 11.09M & xxx & 68.9 & 8.71 & 46.1 & 71.3  & 2.47 \\
 &{\FedAdp}  & 3.19M   & 61.6 & 8.19 & 42.1 & 65.8 & 2.12 \\
 &{\lora}    & 0.49M   & 68.4 & 8.65 & 46.1 & 70.3 & 2.45 \\
 &{\name}    & 0.37M   & 69.4 & 8.77 & 46.9 & 70.9 & 2.43 \\
 \hline
 \multirow{5}{1.3cm}{GPT-2 L (Ideal)} 
 &{FT}       & 774.03M & 68.5 & 8.78 & 46.0 & 69.9 & 2.45 \\
 % &\textit{Adapter_\text{L}}   & 23.00M & xxx & 68.9 & 8.70 & 46.1 & 71.3  & 2.45 \\
 &{\FedAdp}  & 6.05M & 65.7 & 8.53 & 42.2 & 67.9 & 2.08 \\
 &{\lora}    & 1.06M & 68.9 & 8.69 & 46.4 & 70.8 & 2.44 \\
 &{\name}    & 1.06M & 69.8 & 8.82 & 46.8 & 70.7 & 2.51 \\
 \hline
 \multirow{5}{1.3cm}{GPT-2 L (Hetero)} 
 &{FT}       & 774.03M & 60.2 & 8.10 & 40.8 & 64.1 & 2.01 \\
 % &\textit{Adapter_\text{L}}   & 23.00M & xxx & 68.9 & 8.70 & 46.1 & 71.3  & 2.45 \\
 &{\FedAdp}  & 6.05M & 60.3 & 8.09 & 40.9 & 64.0 & 1.90 \\
 &{\lora}    & 1.06M & 67.9 & 8.57 & 45.9 & 70.4 & 2.43 \\
 &{\name}    & 0.79M & 69.6 & 8.80 & 46.3 & 70.6 & 2.49 \\
 \hline
\end{tabular}
}
\caption{GPT-2 medium (M) and large (L) with different adaptation methods on the E2E NLG Challenge.}
\label{tab:GPT2_performance}
\vspace{-3ex}
\end{table*}

\begin{figure*}[t]
  \centering
  \subfloat[Important weights identification\label{fig:hetero_weights}]{
    \includegraphics[width=0.27\linewidth]{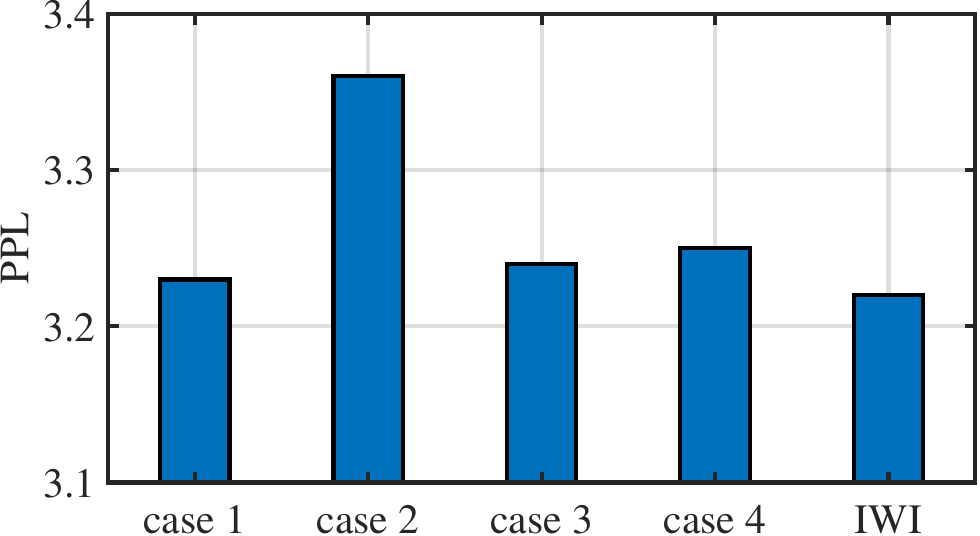}
  }
  \subfloat[Heterogeneous configuration\label{fig:hetero_avg_var_gpt2}]
  {
    \includegraphics[width=0.27\linewidth]{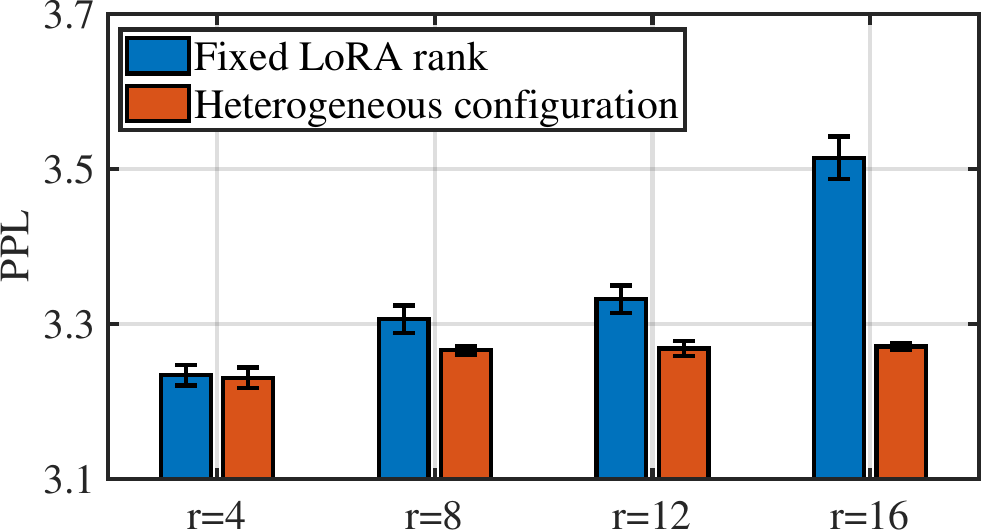}
  }
    \subfloat[Quantization\label{fig:GPT2_lg_quantization}]
  {
    \includegraphics[width=0.29\linewidth]{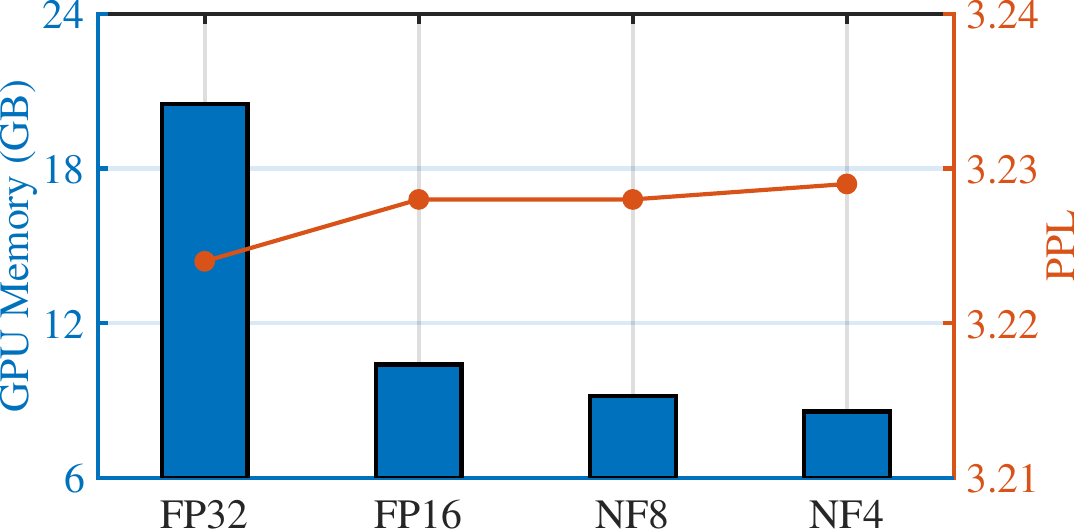}
  }
  % \vspace{-3mm}
  \caption{Ablation Evaluation on GPT2-L model.}
  \label{fig:ppl_gpt2}
  \vspace{-2ex}
\end{figure*}

\subsection{Ablation Evaluation}

\rev{
\quad{\bf{Important weights identification (IWI). }} Fig.~\ref{fig:hetero_weights} shows the impact of the selection of trainable weights on GPT2-L model under \rev{heterogeneous} setting. Considering that the average number of trainable weights in our training process is approximately 3, we compare the IWI method with a random combination of three trainable weights for fair comparison. The proposed IWI method exhibits lower PPL than all three random combinations of weights, which demonstrates the effectiveness of our IWI method. This is because large singular values usually correspond to the primary directions of variation in model weights, reflecting where the density of principal information is high. Therefore, it is efficient to prioritize fine-tuning the trainable weights with large singular values. Additionally, training performance on other metrics is summarized in Table~\ref{tab:iwi}.}

\begin{table}[t]
\centering
\scalebox{0.8}{
\begin{tabular}{ |c||c|c|c|c|c| }
    \hline
    \multirow{2}{*}{Weight Index} &\multicolumn{5}{c|}{E2E NLG Challenge} \\ &BLEU &NIST &MET &ROUGE-L &CIDEr \\
    \hline
    \textit{$W_q+W_k+W_v$} (case 1)   & 69.1 & 8.70 & 46.7 & 70.4 & 2.44 \\
    \textit{$W_q+W_k+W_o$} (case 2)   & 68.1 & 8.76 & 45.3 & 68.9 & 2.39 \\
    \textit{$W_q+W_v+W_o$} (case 3)   & 69.2 & 8.72 & 46.6 & 70.5 & 2.43 \\
    \textit{$W_k+W_v+W_o$} (case 4)   & 69.4 & 8.73 & 46.7 & 70.4 & 2.43 \\
    {IWI}         & 69.6    & 8.80    & 46.3    & 70.6    & 2.49 \\
    \hline
\end{tabular}
}

\caption{\needrev{GPT-2 large model with different combinations of trainable weights on the E2E NLG challenge.}}
\label{tab:iwi}
\vspace{-2ex}
\end{table}

\rev{{\bf{Heterogeneous configuration (HC).}} Fig.~\ref{fig:hetero_avg_var_gpt2} presents the impact of the selection of rank size on model training. We compare our proposed HC method with conventional LoRA schemes with different fixed rank sizes. It is observed that as the rank increases from 4 to 16, the PPL of the conventional LoRA framework grows rapidly from 3.23 to 3.51, nearly 9\%, indicating an increasingly severe influence of the heterogeneous computing resource across clients on model training. This is attributed to the observation that the larger rank implies heavier computing  workloads, leading to lower under-training rates. Conversely, the PPL of the HC method increases slightly with the rank, owing to our design of adaptively adjusting the rank based on heterogeneous computing capabilities. Moreover, the performance gap between HC method and LoRA with fixed rank sizes significantly widens with the increase in rank, further revealing the effectiveness of HC method. Similarly, the training performance on other metrics is shown in Table~\ref{tab:hc}.}

\begin{table}[t]
\centering
\scalebox{0.8}{
\begin{tabular}{ |c||c||c|c|c|c|c| }
    \hline
    {}{} &\multirow{2}{*}{Rank} &\multicolumn{5}{c|}{E2E NLG Challenge} \\ & &BLEU &NIST &MET &ROUGE-L &CIDEr \\
    \hline
    % unconstrained & \textit{r = 4} & 69.6   & 8.79   & 46.4   & 71.5  & 2.51 \\
    % \hline
    \multirow{3}{1.1cm}{Fixed LoRA rank} 
    & \textit{r = 8}     & 66.8    & 8.48    & 46.2    & 70.7    & 2.39 \\
    & \textit{r = 12}    & 68.8    & 8.68    & 46.4    & 70.8    & 2.43 \\
    & \textit{r = 16}    & 69.4    & 8.75    & 46.5    & 71.3    & 2.46 \\
    \hline
    \multirow{3}{1.1cm}{HC} 
    & \textit{$r_{max}$ = 8}  & 69.4    & 8.73    & 46.8    & 71.1    & 2.46 \\
    & \textit{$r_{max}$ = 12} & 69.6    & 8.77    & 46.9    & 70.9    & 2.43 \\
    & \textit{$r_{max}$ = 16} & 69.6    & 8.80    & 46.3    & 70.6    & 2.49 \\
    \hline
\end{tabular}
}
\caption{GPT-2 large model with different heterogeneous configurations on the E2E NLG challenge.}
\label{tab:hc}
\vspace{-2ex}
\end{table}

\rev{{\bf{Quantization.}} Fig.~\ref{fig:GPT2_lg_quantization} illustrates the GPU memory space and training accuracy under different quantization bits. The GPU memory space of FP32 is significantly larger than other quantization levels, 20.5\!~GB, \needrev{where the model takes up 3.14\!~GB,} which is approximately \needrev{1.9, 3.2, and 4.9 times} that of FP16, NF8, and NF4, respectively. Quantization bit makes trade off between memory space and training accuracy of LLMs. It is observed that as the number of quantization bits diminishes, the training accuracy of the model only slightly decreases. FP32 achieves slightly higher training accuracy, i.e., with PPL \needrev{3.224}, whereas FP16, NF8, and NF4 exhibit similar lower training accuracy, with PPL approximately \needrev{3.228}. The proposed quantization scheme further reduces the memory requirements of the \name framework, thereby improving its scalability. A more comprehensive comparison is illustrated in Table~\ref{tab:quantization}.
}

\begin{table} [t]
\centering
\scalebox{0.9}{
\begin{tabular}{ |c||c|c|c|c|c| }
    \hline
    \multirow{2}{*}{Quantization} &\multicolumn{5}{c|}{E2E NLG Challenge} \\ &BLEU &NIST &MET &ROUGE-L &CIDEr \\
    \hline
    {FP32}   & 70.1   & 8.83  & 46.5  & 71.7  & 2.50 \\
    {FP16}   & 69.6   & 8.79  & 46.2  & 70.6  & 2.49 \\
    {NF8}   & 69.5   & 8.75  & 46.8  & 71.0  & 2.44 \\
    {NF4}   & 69.4   & 8.77  & 46.9  & 70.9  & 2.43 \\
    \hline
\end{tabular}
}
\caption{\name on GPT-2 large model with different quantization bits on the E2E NLG challenge.}
\label{tab:quantization}
\vspace{-2ex}
\end{table}

\section{Related Work} \label{sec:rw}

\quad {\bf{Parameter-efficient fine-tuning: }} PEFT has recently emerged as a promising approach for efficient LLM fine-tuning, which strikes a balance between the number of trainable parameters and training accuracy in NLP tasks. Among various PEFT approaches, ADAPTER~\cite{houlsby2019parameter, pfeiffer2020adapterfusion, karimi2021compacter} and LoRA~\cite{hu2021lora} stand out as the most attractive techniques for substantially reducing trainable parameters without compromising LLM fine-tuning performance. ADAPTER inserts small trainable modules into each transformer block while keeping the parameters of the pre-trained model frozen to improve training efficiency. Compacter~\cite{karimi2021compacter} is a variant of ADAPTER that achieves lower parameter complexity based on shared information across adapters and low-rank subspaces of the model. However, the ADAPTER approaches introduce additional computing overhead in adapter layers, increasing the LLM fine-tuning latency, especially with smaller batch sizes. In contrast, LoRA~\cite{hu2021lora} optimizes the variation of dense layers with low-rank decomposition matrices while keeping pre-trained model parameters frozen. Unlike ADAPTER, the low-rank matrices are inserted into the fine-tuned modules in a parallel manner, mitigating any additional training latency and thus enhancing the efficiency of LLM fine-tuning.

{{\bf{FL paradigm for LLMs: }}
PEFT needs to fine-tune the LLMs based on domain-specific data of downstream tasks. However, clients may be reluctant to share raw data with the server due to data privacy concerns, especially privacy-sensitive data such as medical images. A promising solution to this issue is the FL paradigm, which enables multiple users to train models without sharing raw data. Recently, some research efforts have been made to merge LLM fine-tuning with the FL paradigm. ~\cite{lin2022fednlp} is the first work to integrate FL with LLM fine-tuning, comprehensively validating the effectiveness of the FedNLP framework on four
common formulations of NLP tasks: text classification, sequence tagging, question answering, and seq2seq generation. FedAdapter~\cite{cai2023efficient} addresses the high training cost of FedNLP, which expedites the model convergence rate by progressively upgrading adapter configurations as well as continuously profiling future adapter configurations. However, existing works only focus on the traditional language models such as BART or BERT, whereas the most prevalent LLMs (e.g., GPT-2~\cite{radford2019language}) with a significant number of parameters are ignored.
}

{{\bf{Model compression and quantization: }}
The unprecedented number of parameters for LLMs renders computation and storage significant bottlenecks for LLM fine-tuning. Model compression and quantization have emerged as promising techniques to reduce model storage size and accelerate inference by decreasing bit precision~\cite{bernstein2018signsgd, wu2018error}. GPTQ~\cite{frantar2022gptq} leverages second-order information for error compensation, significantly enhancing model accuracy and training efficiency. 
\needrev{The authors of~\cite{lin2023awq} propose an edge-cutting quantization method, named AWQ, to quantize weights into low-bit integers, thereby reducing the hardware barrier and speeding up token generation.} 
Unlike previous work on quantization for LLM inference, SwitchBack layers~\cite{wortsman2023stable} investigates backpropagation through quantized weights over 1B parameters and QLoRA~\cite{dettmers2023qlora} quantizes weights into 4-bit NormalFloat and backpropagates gradients into low-rank structure with the pre-trained LLMs.
}

{{\bf{Resource heterogeneity: }}
The resource heterogeneity~\cite{horvath2021fjord,lin2023fedsn, li2022pyramidfl} poses a significant challenge to deploying FL frameworks~\cite{wang2021device, tran2019federated, wang2018wide} for LLMs. Clients with limited computing or memory resources require more time to complete local model updates, impeding the FL training process or even being excluded from model training. Although there are papers considering resource heterogeneity in general FL \cite{zhu2024esfl}, there is no existing work that addresses resource heterogeneity in FL frameworks for LLM fine-tuning.
}

\section{Conclusion} \label{sec:con}
In this paper, we have proposed and implemented \name, an automated federated pipeline, to facilitate LLM FL fine-tuning in edge servers with heterogeneous computing resources. We have modeled the problem as a MILP to guide our design. To solve the problem, we have first selected important weights, and configured the corresponding parameters under computing budgets at edge servers for their LoRA adapters. We have also quantized LoRA adapters to further reduce \rev{memory} space. Our extensive evaluations have demonstrated that \name achieves significantly better performance than the state-of-the-art baselines in LLM FL.

\ifCLASSOPTIONcaptionsoff
  \newpage
\fi

\bibliographystyle{IEEEtran}
\bibliography{reference}

\end{document}